\documentclass[10pt]{article}
\usepackage[legalpaper, margin=1in]{geometry}

\usepackage{graphicx}

\usepackage{tikz}
\usepackage{comment}
\usepackage{amsmath,amssymb} % define this before the line numbering.
\usepackage{wrapfig}
\usepackage{hyperref}
\usepackage{amsthm}
\usepackage[accsupp]{axessibility}  % Improves PDF readability for those with disabilities.
\usepackage{booktabs}
\usepackage{enumitem}
\usepackage{subcaption}
\usepackage{bbm}
\usepackage{multirow}

\usepackage{times}
\usepackage{epsfig}
\usepackage{tcolorbox}
\usepackage{url}
\usepackage{ulem}
\usepackage{nicefrac} % compact symbols for 1/2, etc.
\usepackage{microtype} % microtypography
\usepackage{xcolor} % colors
\usepackage[ruled,vlined,algo2e]{algorithm2e}
\usepackage{algorithm}

\usepackage{caption}
\usepackage{xr}

\usepackage[T1]{fontenc}

\usepackage{authblk}

\newtheorem{definition}{Definition}

\newtheorem{theorem}{Theorem}

\newtheorem{lemma}{Lemma}

\def\bx{\mathbf{x}}

\begin{document}

\title{A Theoretical Perspective on Subnetwork Contributions to Adversarial Robustness} 
\author[1]{Jovon Craig}
\author[1]{Josh Andle}
\author[2]{Theodore S. Nowak}
\author[1,*]{Salimeh Yasaei Sekeh}
\affil[1]{University of Maine}
\affil[2]{Pacific Northwest National Lab}
\affil[*]{Corresponding Author: salimeh.yasaei@maine.edu}
\setcounter{Maxaffil}{0}
\renewcommand\Affilfont{\itshape\small}

\date{}

\maketitle

\begin{abstract}
The robustness of deep neural networks (DNNs) against adversarial attacks has been studied extensively in hopes of both better understanding how deep learning models converge and in order to ensure the security of these models in safety-critical applications. Adversarial training is one approach to strengthening DNNs against adversarial attacks, and has been shown to offer a means for doing so at the cost of applying computationally expensive training methods to the entire model. To better understand these attacks and facilitate more efficient adversarial training, in this paper we develop a novel theoretical framework that investigates how the adversarial robustness of a subnetwork contributes to the robustness of the entire network. To do so we first introduce the concept of {\it semirobustness}, which is a measure of the adversarial robustness of a subnetwork. Building on this concept, we then provide a theoretical analysis to show that {\it if a subnetwork is semirobust and there is a sufficient dependency between it and each subsequent layer in the network, then the remaining layers are also guaranteed to be robust}. 
We validate these findings empirically across multiple DNN architectures, datasets, and adversarial attacks. Experiments show the ability of a robust subnetwork to promote full-network robustness, and investigate the layer-wise dependencies required for this full-network robustness to be achieved.
\end{abstract}

%%%%%%%%% BODY TEXT
\section{Introduction}
Deep neural networks (DNNs) have been highly successful in computer vision, speech recognition, and natural language processing applications where they have been shown to surpass human performance~\cite{mnih2015human, radford2015unsupervised, goodfellow2016deep}. Despite these successes, the reliability of deep learning algorithms in real-world applications is fundamentally challenged by the existence of adversarial attacks and their generated ``adversarial examples'', which are inputs perturbed with noise such that they fool normally performant networks. In the context of image classification, these perturbations can be unnoticeable to the naked eye yet capable of changing the label of a normally correctly-classified image~\cite{szegedy2014intriguing, goodfellow2014explaining}. For this reason, adversarial examples present a major threat to the security of deep-learning systems. Classifiers can be made robust however, such that they can correctly label adversarially perturbed images. For example, an adversary could alter images of the road to fool a self-driving car's neural network into misclassifying traffic signs \cite{papernot2016limitations}, reducing the car's safety, whereas a robust network would detect and reject the adversarial inputs~\cite{ma2018characterizing, biggio2013evasion}. The methods which generate these adversarial attacks have been studied extensively~\cite{DBLP:Kurakin2017, sharif2016accessorize, brown2017adversarial, eykholt2018robust}, as have the methods for defending against them. These defenses primarily fall under two categories: (1) Efficient methods to detect adversarial examples~\cite{su2019one, laidlaw2019functional, athalye2018synthesizing, liu2016delving, xie2017adversarial, akhtar2018threat}, (2) Adversarial training to make deep neural networks more robust against adversarial attacks~\cite{madry2018towards, tsipras2019robustness, gilmer2019adversarial, ilyas2019adversarial, papernot2016distillation}.

Adversarial perturbations may be applied to the input or to the network's hidden layers \cite{szegedy2014intriguing, goodfellow2014explaining}, and it has been show that this strategy is effective at improving a network's robustness \cite{goodfellow2014explaining}. Several theories have been developed to explain the phenomenon of adversarial examples~\cite{ma2018characterizing,raghunathan2018certified, xiao2019training, cohen2019certified, shamir2019simple, fawzi2016robustness,carlini2017towards,weng2018evaluating}.
Previously \cite{ilyas2019adversarial} sought to develop a theoretical framework for adversarial robustness. The authors categorize a given feature $f$ as ``$\rho$-useful'' and ``$\gamma$-robustly useful''. Those features which are highly-predictive and correlated with the true label in expectation are ``$\rho$-useful''. Those which remain highly-predictive under adversarial attack are said to be ``$\gamma$-robustly useful'' and those which don't are described as ``useful, non-robust features''. The authors postulate that these ``useful, non-robust features'' are responsible for model vulnerability. While \cite{ilyas2019adversarial} introduces a valuable mechanism with which to frame adversarial attacks, neither it nor its proceeding works have investigated how adversarial robustness can be distributed across subnetworks.

In this paper, we develop a new theoretical framework that studies the robustness of subnetworks in a DNN and explains that if the early layers of a network are adversarially trained to be robust, and are {\it sufficiently connected} with the rest of the network, then the robustness of later layers and the entirety of the network can be guaranteed. 
In doing so, we shed light on a mechanism by which to reduce the computational complexity of adversarial training methods ~\cite{zhang2019you}, as well as the potential need for more training data and memory capacity~\cite{schmidt2018adversarially}. To this end we analyze the conditions under which the adversarial robustness of the full network can be guaranteed given the robustness of a subnetwork by adversarially training only a subset of layers. This includes the introduction of the concept of ``semirobustness'' as a means with which to measure the adversarial robustness of a subnetwork, and a novel theoretical framework that explains theories for the following claim:

\begin{center}
\begin{tcolorbox}
% \vspace{-.06in}
%\emph{
\textit{If a subnetwork is robust and there is sufficient dependency between the subnetwork and subsequent layers, then the remaining layers are also guaranteed to be robust.}

% \vspace{-0.05in}
\end{tcolorbox}
\end{center}

\paragraph{Contributions} In summary, our contributions are: (1) We {\bf introduce the novel concept of ``semirobustness''} for subnetworks. We show that a subnetwork is semirobust if and only if all layers within it are semirobust. (2) We {\bf provide a novel  theoretical framework} and prove that, under some assumptions, if an early subnetwork is semirobust then the proceeding subnetworks are robust. (3) We empirically investigate the extent of mutual dependency between subnetworks which is sufficient to achieve full network robustness.%, our method displays the same adversarial robustness of a network as compared to regular adversarial training.

\section{Subnetwork Robustness}
\paragraph{Notations}
We assume that a given DNN has a total of $n$ layers, where $F^{(n)}$ is a function mapping the input space $\mathcal{X}$ to a set of classes $\mathcal{Y}$, i.e. $F^{(n)}: \mathcal{X}\mapsto \mathcal{Y}$ . We define $f^{(l)}$ as the $l$-th layer of $F^{(n)}$; $F^{(i,j)}:= f^{(j)}\circ\ldots \circ f^{(i)}$ is a subnetwork, which is a group of consecutive layers $f^{(i)},\ldots,f^{(j)}$. We define $F^{(j)}:= F^{(1,j)}=f^{(j)}\circ\ldots \circ f^{(1)}$ as the first part of the network up to layer $j$. We denote $\sigma ^{(l)}$ the activation function in layer $l$ and $\pi(y)$ the prior probability of class label $y\in \mathcal{Y}$. Let $f^{(l)}$ be the $l$-th layer of $F^{(n)}$, as $f^{(l)}(x_{l-1}) = \sigma^{(l)}(w^{(l)}x_{l-1} + b^{(l)})$, where $\sigma ^{(l)}$ is the activation function. In this section, we define a {\it semirobust subnetwork}. We discuss semirobustness further in Section~\ref{section:theory}.

\begin{definition}\label{def.semirobust}
{\rm (Semirobust Subnetwork)} Suppose input $\mathbf{X}$ and label $y$ are samples from joint distribution $\mathcal{D}$. For a given distribution $\mathcal{D}$, a subnetwork $F^{(j)}$ is called {\it {$\gamma_j$}-semirobust} if there exists a mapping function {$G_j: \mathcal{L}_j \mapsto \mathcal{Y}$} such that
\begin{equation}\label{semi-def:fromula}
 {\mathbb{E}}_{(\mathbf{X},y)\sim \mathcal{D}}\big[\inf_{\delta\in S_x} y \cdot {{G_j}}\circ F^{(j)}(\mathbf{X}+\delta)\big]\geq {{\gamma_j}},
\end{equation}
for an appropriately defined set of perturbations $S_x$. In (\ref{semi-def:fromula}), $G_j$ is a non-unique function mapping layer $f^{(j)}$ to class set $\mathcal{Y}$, and $\gamma_j$ is a constant denoting the correlation between $y$ and $F^{(j)}$.
\end{definition}

Note that $G_j$ is necessary if the dimensionality of $F^{(j)}$ does not match that of $y$, but if $F^{(j)} = F^{(n)}$, the semirobust definition becomes standard $\gamma$-robustness as defined in \cite{ilyas2019adversarial}. { To define semirobustness for a single layer $f^{(j)}$, in (\ref{semi-def:fromula}) we simply replace $f^{(j)}$ in $F^{(j)}$ and $K_{j-1}\circ (\mathbf{X}+\delta)$ in $\mathbf{X}+\delta$, where $K_{j-1}$ is mapping function $K_{j-1}: \mathcal{X}\mapsto\mathcal{L}_{j-1}$. To avoid confusion, in this paper we use $\mathbf{X}+\delta$ for layer semirobustness as an input as well.}
Throughout this paper, we assume that the network $F^{(n)}$ is a useful network i.e. for a given  distribution $\mathcal{D}$, the correlation between $F^{(n)}$ and true label $y$, ${\mathbb{E}}_{(\mathbf{X},y)\sim \mathcal{D}}\big[y\cdot F^{(n)}(\mathbf{X})\big]$ is highest in expectation in optimal performance. Intuitively, a highly useful network $F^{(n)}$ minimizes the classification loss ${\mathbb{E}}_{(\mathbf{X},y)\sim \mathcal{D}}\big[\mathcal{L}(\mathbf{X},y)\big]$ that is 
\begin{equation}\label{loss-function}
- {\mathbb{E}}_{(\mathbf{X},y)\sim \mathcal{D}}\Big[y\cdot\big(b+\sum_{F^{(n)}\in\mathcal{F}^{(n)}}w_{F^{(n)}} F^{(n)}(\mathbf{X})\big)\Big], 
\end{equation}
where $w_{F^{(n)}}$ is the weight vector and $\mathcal{F}^{(n)}$ is the set of $n$-th layer networks. 
Definition~\ref{def.semirobust} raises valid questions regarding the relationship between a subnetwork and its associated layers' robustness. We show this relationship under the following thoerem.
\begin{theorem}\label{thm.0}
The subnetwork $F^{(j)}$ is $\gamma_j$-semirobust if and only if every layer of $F^{(j)}$, i.e. $f^{(j)},f^{(j-1)},\ldots,f^{(1)}$, is also semirobust with bound parameters $\gamma_j,\ldots,\gamma_1$ respectively. 
\end{theorem}
Theorem~\ref{thm.0} is a key point used to support our main claims on the relationship between layer-wise and subnetwork robustness, and its proof is provided as supplementary materials (SM).
In the next section, we show that the robustness of subnetworks are guaranteed under a strong dependency assumption between layers.
\subsection{Semirobustness guarantees}\label{section:theory}
In this section, we provide theoretical analysis to explain how the dependency between the layers of subnetworks promotes semirobustness and eliminates the requirement of adversarially training the entire network.  
\paragraph{Non-linear Probabilistic Dependency (Mutual Information):} Among other various probabilistic dependency measures, in this paper, we adopt an information-theoretic measure called mutual information (MI): a measure of the reduction in uncertainty about one random variable by knowing about another. Formally, it is defined as follows: Let $\mathcal{X}$ and $\mathcal{Z}$ be Euclidean spaces, and let $P_{XZ}$ be a probability measure in the space $\mathcal{X}\times \mathcal{Z}$. 
Here, $P_X$ and $P_Z$ define the marginal probability measures. The mutual information (MI), denoted by $I(X;Z)$, is thus defined as, 
\begin{equation}\label{def:MI}
  I(X;Y) = \mathop{\mathbb{E}}_{P_X P_Z}\left[  g\left(\frac{dP_{XZ}}{dP_X P_Z}\right)\right], 
\end{equation}
where $\frac{dP_{XZ}}{dP_X P_Z}$ is the Radon-Nikodym derivative, $g:(0,\infty)\mapsto \mathbb{R}$ is a convex function, and $g(1)=0$.
Note that when $\frac{dP_{XY}}{dP_X P_Y}\rightarrow 1$, then $I\rightarrow 0$. Using (\ref{def:MI}), the MI measure between two layers $f^{(i)}$ and $f^{(j)}$ with joint distribution $P_{ij}$ and marginal distributions $P_i$, $P_j$ respectively is given as
\begin{equation}\label{def:Network-MI}
  I(f^{(i)};f^{(j)}) = \mathop{\mathbb{E}}_{P_i P_j}\left[  g\left(\frac{dP_{ij}}{dP_i P_j}\right)\right].  
\end{equation}
The concept of MI is integral to the most important theory in our theoretical framework through the assumptions below.\\
{\bf Assumptions:} Let $G_a:\mathcal{L}_a \mapsto \mathcal{Y}$ be a function mapping layer $f^{(a)}$ to a label $y\in\mathcal{Y}$, and let $G_j:\mathcal{L}_j \mapsto \mathcal{Y}$ be a function mapping layer $f^{(j)}$ to a label $y\in\mathcal{Y}$. Let $g_\delta = f^{(a)}(\mathbf{X}+\delta)$ and $h_{\delta,j} = f^{(j)}(\mathbf{X}+\delta)$ for $\delta\in S_x$ (perturbation set). Note that $g_\delta=h_{\delta,a}$.\\
 {\bf A1}: The class-conditional MI between $h_{\delta,j-1}$ and $h_{\delta,j}$ is at least hyperparameter $\rho_j\geq 0$, i.e. 
    \begin{align}\label{assumption:A1}
    \sum\limits_y \pi(y) I\left(h_{\delta,j-1};h_{\delta,j}|y\right) \geq \rho_j
    \end{align}
 {\bf A2}: There exists a constant $U_j\geq 0$ such that for all $\delta\in S$:
\begin{align*}
\mathbb{E}_{p(h_{\delta,j-1}, h_{\delta,j},y)}\left[\frac{p(h_{\delta,j-1}, h_{\delta,j}|y)}{p(h_{\delta,j-1}|y)p(h_{\delta,j}|y)} \right]\leq U_j,\;\;\;\; \hbox{and}\;\;\;\;\\[5pt]
    \mathbb{E}_{p(h_{\delta,j-1}, h_{\delta,j},y)}\left[y \cdot (G_j \circ h_{\delta,j} - G_{j-1} \circ h_{\delta,j-1})\right]\geq 1+U_j
\end{align*}
where $p(h_{\delta,j-1}, h_{\delta,j},y)$ is the joint probability of random triple $(h_{\delta,j-1}, h_{\delta,j},y)$.  
\begin{theorem}\label{thm.1}
Let $f_a$ be a $\gamma_a$-semirobust subnetwork equivalent to $F^{(a)}$, and let $f_b$ be the subnetwork $F^{(a+1,n)}$ and for $j=a+1,\ldots,n$, assumptions {\bf A1} and {\bf A2} holds true. 
Then $f_b$ is $\gamma_b$-semirobust.
\end{theorem}
In Theorem~\ref{thm.1}, $\gamma_b\leq\gamma_a+\sum_{j=a+1}^b\rho_j$. Note that the constant $U_j$ does not depend on $\gamma_a$, $\gamma_b$, and $\rho_j$. This theorem is an extension of the following lemma, and the proofs of both are found in the SM.
\begin{lemma}\label{lemma.MI}
Let $F^{(n-1)}$ be a $\gamma_{n-1}$-semirobust subnetwork. Let $g_\delta = f^{(n-1)}(\mathbf{X}+\delta)$ and $h_\delta = f^{(n)}(\mathbf{X}+\delta)$ for $\delta\in S_x$. Let $G_{n-1}:\mathcal{L}_{n-1}\mapsto \mathcal{Y}$ be a function mapping layer $g$ to the network's output $y\in\mathcal{Y}$. Under the following assumptions $f^{(n)}$ is $\gamma_n$-semirobust:\\
 {\bf B1}: The MI between $f^{(n-1)}$ and $f^{(n)}$ is at least hyperparameter $\rho\geq 0$, i.e.
\begin{align}\label{assumption:B1}
 \sum\limits_y \pi(y) I\left(g_\delta;h_\delta|y\right) \geq \rho.
\end{align}
{\bf B2}: There exists a constant $U\geq 0$ such that for all $\delta\in S$:
\begin{align*}\mathbb{E}_{p(g_\delta, h_\delta,y)}\left[\frac{p(g_\delta, h_\delta|y)}{p(g_\delta|y)p(h_\delta|y)} \right]\leq U, \;\;\;\hbox{and}\\[5pt]
\mathbb{E}_{p(g_\delta, h_\delta,y)}\left[y \cdot (h_\delta -G_{n-1} \circ g_\delta)\right]\geq 1+U.
\end{align*}
\end{lemma}
{Note that in Lemma~\ref{lemma.MI}, $\gamma_{n}\leq \gamma_{n-1}+\rho$}, and assumptions ${\bf B1}$ and ${\bf B2}$ are particular cases of ${\bf A1}$ and ${\bf A2}$, when $a=n-1$.\\
{\bf Intuition:} Let $\mathcal{IF}(.)$ determine the information flow passing through layers in the network $F^{(n)}$. Intuitions from the $\mathcal{IF}$ literature would advocate that in a feed-forward network if the learning information is preserved up to a given layer, one can utilize knowledge of this information flow in the next consecutive layer's learning process due to principle $F^{(i,j)}=f^{(j)}\circ F^{(i,j-1)}$, and consequently $\mathcal{IF}^{(i,j)}\approx\mathcal{IF}^{(j)}\circ \mathcal{IF}^{(i,j-1)}$. This is desirable as in practice training the subnetwork requires less computation and memory usage. This explains that under the assumption of the strong connection between $j$-th and $j-1$-th layers, the information automatically passes throughout the later layers, and subnetwork training returns sufficient solutions for task decision-making. To better characterize the measure of information flow, we employ a non-linear and probabilistic dependency measure that determines the mutual relationship between layers and how much one layer tells us about the other one. An important takeaway from Theorem~\ref{thm.1} (and Lemma~\ref{lemma.MI}) is that a strong non-linear mutual connectivity between subnetworks guarantees that securing only the robustness of the first subnetwork ensures information flow throughout the entire network.
\paragraph{Linear Connectivity:} To provably show that our theoretical study in Theorem~\ref{thm.1} is satisfied for the linear connectivity assumption between subnetworks, we provide a theory that investigates the scenario when the layers in the second half of the network are a linear combination of the layers in the first subnetwork.  
\begin{theorem}\label{thm.2}
Let $f_a$ be a $\gamma_a$-semirobust subnetwork equivalent to $F^{(a)}$, and let $f_b$ be the subnetwork $F^{(a+1,n)}$. If for $j=a+1,\ldots,n$, $f^{(j)}=\sum\limits_{i=1}^{j-1}\lambda_{ij}^{T}.f^{(i)}$, where $\lambda_{ij}$ is a map $\mathcal{L}_i \mapsto \mathcal{L}_j$ and a matrix of dimensionality $\mathcal{L}_i\times \mathcal{L}_j$, then $f_b$ is $\gamma_b$-semirobust where $\gamma_b=\gamma_a\big((n-1-a)(n-a)\big/2\big)$.
\end{theorem}
This theorem shows that when the connectivity between layers in $f_a$ and $f_b$ is linear, we achieve the semirobustness property for the subnetwork $f_b$. Importantly, note that linear combination multipliers determine the Pearson correlation between layers given the constant variance of the layers. This is because if $f^{(j)}=\lambda_{ij}f^{(i)}$, then $Corr(f^{(j)},f^{(i)})=\lambda_{ij}var(f^{(i)})$. Theorem~\ref{thm.2} is an extension of the lemma~\ref{lemma.Lin.Comb}. Detailed proof and accompanying experiments are provided in the SM. 
\begin{lemma} \label{lemma.Lin.Comb}
Let the last layer $f^{(n)}$ be a linear combination of $f^{(n-1)},\ldots,f^{(1)}$, expressed as $f^{(n)}=\sum\limits_{i=1}^{n-1}\lambda_{i}^{T} \cdot f^{(i)}$, where $\lambda_i$ is a map $\mathcal{L}_i \mapsto \mathcal{L}_n$ and a matrix of dimensionality $\mathcal{L}_i\times \mathcal{L}_n$. If $F^{(n-1)}$ is $\gamma$-semirobust, then $f^{(n)}$ is $\gamma_n$-semirobust where $\gamma_n=\sum\limits_{i=1}^{n-1}\gamma_i$.
\end{lemma}
{\bf Question:} At this point, a valid argument could be how the performance of a network differs under optimal full-network robustness, $(f^*_a,f^*_b)$ and subnetwork robustness $(f^*_a,\widetilde{f}_b)$. Does the difference between performance have any relationship with the weight difference of subnetworks $f^*_b$ and $\widetilde{f}_b$? This question is investigated in the next section by analyzing the difference between loss function of the networks $(f^*_a,f^*_b)$ and $(f^*_a,\widetilde{f}_b)$.
\subsection{Further Theoretical Insights}
Let $\omega^*$  be the convergent parameters after training has been finished for the network $F^{*(n)}:=(f^*_a,f^*_b)$, that is adversarially robust against a given attack. Let $\widetilde{\omega}^*$ be the convergent parameters for network $(f_a^*, \widetilde{f_b})$, that is adversarially semirobust against the attack. This means that only the first half of the network is robust against attacks. Let ${\omega}^*_b$, $\widetilde{\omega}_b$, and ${\omega}^*_a$ be weights of networks $f^*_b$, $\widetilde{f}_b$, and $f^*_a$, respectively. Recall the loss function (\ref{loss-function}), and remove offset $b$ without loss of generality. \begin{equation}\label{loss-func}
\hbox{Define}\;\;\;\;\ell(\omega):=-\sum\limits_{F\in\mathcal{F}}w_{F}\cdot F^{(n)}(\mathbf{X}),
\end{equation}
therefore the loss function in (\ref{loss-function}) becomes 
\begin{align*}    
\mathbb{E}_{(\mathbf{X},Y)\sim D}\{L(F^{(n)}(\mathbf{X}),Y)\} =\mathbb{E}_{(\mathbf{X},Y)\sim D}\left\{Y\cdot \ell(\omega)\right\}, \;\hbox{and}\;\; 
\end{align*}
\begin{align}
      \omega^*:= {argmin}_{\omega} \mathbb{E}_{(\mathbf{X},Y)\sim D}\left\{ Y\cdot\left(\ell({\omega})\right)\right\}, 
\end{align}
where $\ell$ is defined in (\ref{loss-func}).\\
\begin{definition}\label{perform-diff}
{\rm (Performance Difference)} Suppose input $\mathbf{X}$ and task $Y$ have joint distribution $\mathcal{D}$. Let $\widetilde{F}^{(n)}:=(f^*_a,\widetilde{f}_b) \in \mathcal{F}$ be the network with $n$ layers when the subnetwork $f^*_a$ is semirobust. The performance difference between robust $F^{*(n)}:=(f^*_a,f^*_b)$ and semirobust $\widetilde{F}^{(n)}$ is defined as
\begin{equation}\label{training-deviation}
\begin{aligned}
   d(F^{*(n)},\widetilde{F}^{(n)}):= \mathbb{E}_{(\mathbf{X},Y)\sim D}\left\{L(F^{*(n)}(\mathbf{X}),Y) - L(\widetilde{F}^{(n)}(\mathbf{X}),Y) \right\}.
\end{aligned}
\end{equation}
Let ${\delta} (\omega^*|\widetilde{\omega}^*):= \ell(\omega^*)-\ell_t(\widetilde{\omega}^*).$ The performance difference (\ref{training-deviation}) is the average of ${\delta}$:

\begin{equation}\label{optimal-difference}
\begin{aligned}
d(F^{*(n)},\widetilde{F}^{(n)}) &= \mathbb{E}_{(\mathbf{X},Y)\sim D}\left[Y\cdot{\delta} (\omega^*|\widetilde{\omega}^*)\right]\\
&= \mathbb{E}_{(\mathbf{X},Y)\sim D}\big[Y\cdot\big(\ell(\omega^*)-\ell(\widetilde{\omega}^*)\big)\big].
\end{aligned}
\end{equation}

\end{definition}
   Using Taylor approximation of $\ell$ around ${\omega}^*$:
   \begin{equation}
   \begin{aligned}
   \ell(\widetilde{\omega}^*) \approx \ell({\omega}^*) + (\widetilde{\omega}^* - {\omega}^*)^T \nabla\ell({\omega}^*)+ \frac{1}{2}(\widetilde{\omega}^* - {\omega}^*)^T \nabla^2\ell({\omega}^*) (\widetilde{\omega}^* - {\omega}^*),
   \end{aligned}
   \end{equation}
   where $\nabla\ell({\omega}^*)$ and $\nabla^2\ell({\omega}^*)$ are gradient and Hessian for loss $\ell$ at ${\omega}^*$. Since ${\omega}^*$ is the convergent points of $(f^*_a,f^*_b)$, then  $\nabla\ell({\omega}^*)=0$, this implies
\begin{equation}\label{eq-new:1}
\begin{aligned}
 \ell(\widetilde{\omega}^*) - \ell({\omega}^*) &\approx \frac{1}{2}(\widetilde{\omega}^* - {\omega}^*)^T \nabla^2\ell({\omega}^*) (\widetilde{\omega}^* - {\omega}^*)\\ &\leq \frac{1}{2} \lambda^{max} \|\widetilde{\omega}^* - {\omega}^*\|^2,
\end{aligned}
\end{equation}
where $\lambda^{max}$ is the maximum eigenvalue of $\nabla^2\ell({\omega}^*)$. In (\ref{eq-new:1}) we can write $\|\widetilde{\omega}^* - {\omega}^*\|^2=\|\widetilde{\omega}_b - {\omega}_b^*\|^2$ holds because $\widetilde{\omega}^*=(\omega_a^*,\widetilde{\omega}_b)$ and ${\omega}^*=(\omega_a^*,\omega_b^*)$. Note that here the weight matrices $\omega^*$ and $\widetilde{\omega}^*$ are reshaped. Using the loss function $\mathbb{E}_{(\mathbf{X},Y)\sim D}\left\{Y\cdot \ell(\omega)\right\}$, we have
\begin{equation}
\begin{aligned}\label{claim.1}
    \mathbb{E}_{(\mathbf{X},Y)\sim D}\left\{Y\cdot \big(\ell(\widetilde{\omega}^*) - \ell({\omega}^*)\big)\right\} \leq \frac{1}{2}\mathbb{E}_{(\mathbf{X},Y)\sim D}\left\{Y\cdot\big(\lambda^{max} \|\widetilde{\omega}_b - {\omega}_b^*\|^2\big)\right\}. 
\end{aligned}    
\end{equation}
This explains that the performance difference (\ref{optimal-difference}) between networks $F^{*(n)}$ and $\widetilde{F}^{(n)}$ is upper bounded by the $L_2$ norm of weight difference of $f^*_b$ and $\widetilde{f}_b$ i.e. $\widetilde{\omega}_b - {\omega}_b^*$. 

Alternatively, using Cauchy–Schwarz inequality, we have
\begin{equation}\label{eq1:discussion}    
\begin{aligned}
 \mathbb{E}_{(\mathbf{X},Y)\sim D}\left\{Y\cdot \big(\ell(\widetilde{\omega}^*) - \ell({\omega}^*)\big)\right\}\leq \mathbb{E}_{(\mathbf{X},Y)\sim D}\left\{Y\|f^{(n)}(\mathbf{x};\widetilde{\omega}^*)-f^{(n)}(\mathbf{x};{\omega}^*)\|_2\right\},
\end{aligned}
\end{equation}
where $f^{(n)}$ is the last layer of the network. Recall (8) from \cite{lee2021layer}. As $\widetilde{\omega}^*$ and ${\omega}^*$ are the weights of network on $(f^*_a,f^*_b)$ and $(f^*_a,\widetilde{f}_b)$, we have
\begin{equation}    
\begin{aligned}
 \|f^{(n)}(\mathbf{x};\widetilde{\omega}^*)-f^{(n)}(\mathbf{x};{\omega}^*)\|_2 \leq    \|\widetilde{\omega}^*_b - {\omega}^*_b\|_F\|\sigma \left(f_a(\mathbf{x},\omega^*_a)\right)\|_2.
\end{aligned}
\end{equation}
next, we assume the activation function $\sigma$ is Lipschitz continuous i.e. for any $\mathbf{u}$ and $\mathbf{v}$ there exist constant $C^{\sigma}$ s.t. $|\sigma(\mathbf{u})-\sigma(\mathbf{v})|\leq C^{\sigma}|\mathbf{u}-\mathbf{v}|$. Next, assume the activation function is satisfied in $\sigma(\mathbf{0})=\mathbf{0}$. Further by assuming that $\|\mathbf{x}\|_2$ is bounded by $C_x$ and by using peeling procedure, we get:
\begin{equation}\label{eq2:discussion}    
\begin{aligned}
\|f^{(n)}(\mathbf{x};\widetilde{\omega}^*)-f^{(n)}(\mathbf{x};{\omega}^*)\|_2 \leq   C_{\mathbf{x},\sigma} \|\widetilde{\omega}^*_b - {\omega}^*_b\|_F \prod_{j\in a}{\|\omega^{*(j)}\|_F},
\end{aligned}
\end{equation}
here $\omega^{*(j)}$ is the weight matrix of layer $j$-th in $f^*_a$ and $C_{\mathbf{x},\sigma}=C_{\mathbf{x}}C_\sigma$. Combining (\ref{eq1:discussion}) and (\ref{eq2:discussion}) we provide the upper bound:
\begin{equation}\label{eq3:discussion}
\begin{aligned}
 \mathbb{E}_{(\mathbf{X},Y)\sim D}\left\{Y\cdot \big(\ell(\widetilde{\omega}^*) - \ell({\omega}^*)\big)\right\} \leq \mathbb{E}_{(\mathbf{X},Y)\sim D}\big\{Y\cdot\big(C_{\mathbf{x},\sigma}(\omega^*_a) \|\widetilde{\omega}^*_b - {\omega}^*_b\|_F \big)\big\},
\end{aligned}
\end{equation}
where $C_{\mathbf{x},\sigma}(\omega^*_a)=C_{\mathbf{x},\sigma}\prod_{j\in a}{\|\omega^{*(j)}\|_F}$. This alternative approach validates the result shown in (\ref{claim.1}) and aligns with the conclusion that the performance difference between robust and semirobust networks is highly related to their weight differences. In this section we proved two bounds for performance difference defined in (\ref{optimal-difference}).

\section{Experiments and Analysis}
\subsection{Experimental Setup}
To guide the empirical investigation of our theoretical findings, we perform a series of experiments aimed at investigating the necessary conditions for Theorems~\ref{thm.0}-\ref{thm.2}. Here we outline the details of the experiment setup, including attack models, MI estimator, and training settings.
\paragraph{Attack Models:} The most common threat model used when generating adversarial examples is the additive threat model. Let $\mathbf{X}=(X_1,\ldots,X_d)$, where each $X_i\in \mathcal{X}$ is a feature of $\mathbf{X}$. In an additive threat model, we assume adversarial example  $\mathbf{X}_\delta=(X_1+\delta_1,\ldots,X_d+\delta_d)$, i.e. $\mathbf{X}_\delta=\mathbf{X} + \delta$ where $\delta=(\delta_1,\ldots,\delta_d)$. Under this attack model, perceptual similarity is usually enforced by a bound on the norm of $\delta$, $\|\delta\|\leq \epsilon$. %Commonly used norms include $\|.\|_2$, Euclidean distance. 
Note that a small $\epsilon$ is usually necessary because otherwise, the noise on the input could be visible. For evaluating robustness we use some of the most common additive attack models: the Fast Gradient Sign Method (FGSM) \cite{goodfellow2014explaining}, Projected Gradient Descent (PGD) \cite{madry2018towards}, and AutoAttack~\cite{croce2020reliable}. We use $\epsilon=\frac{8}{255}$ for all attacks during training and evaluation. For iterative attacks we use a step size of $\frac{2}{255}$ with 10 iterations. All attacks use an $L_\infty$-norm. The Torchattacks \cite{kim2020torchattacks} library is used for generating all attacks during training and evaluation.

\paragraph{Training Setup:} All experiments are run with ResNet-18 \cite{he2016deep} and WideResNet-34-10 \cite{sergey2016wide} on the CIFAR-10 and CIFAR-100 datasets~\cite{krizhevsky2009learning}. Training (both regular and adversarial) is done with a Stochastic Gradient Descent optimizer for 120 epochs with a learning rate of 0.01 which decreases by a factor of 10 at epochs 40 and 80, using a weight decay of $5\times10^{-4}$ and batch size of 512. 
For adversarial training we use PGD with 10 iterations, $\epsilon=\frac{8}{255}$, and step size of $\frac{2}{255}$ under an $L_\infty$-norm bound.
When denoting the size of a subnetwork such as $f_b$, the number of layers refers to the number of convolutional or linear layers included in the subnetwork. For an $f_b$ of 12 layers, $f_b$ starts at the 12th convolutional or linear layer from the output, including all intermediate (e.g. batch normalization) layers. When training a single subnetwork such as $f_b$, the rest of the network is kept frozen during the training. 

\paragraph{MI Estimation:} We use a reduced-complexity MI estimator called the ensemble dependency graph estimator (EDGE) \cite{noshad2019scalable}. The estimator combines randomized locality-sensitive hashing (LSH), dependency graphs, and ensemble bias-reduction methods. We chose EDGE because it has been shown that it achieves optimal computational complexity $O(n)$, where $n$ is the sample size. It is thus significantly faster than its plug-in competitors \cite{kraskov2004estimating,moon2017ensemble,noshad2017direct}. In addition to fast execution, EDGE has an optimal parametric MSE rate of $O(1/n)$ under a specific condition. 

\subsection{Learning Hyperparameter \texorpdfstring{$\rho$}{Rho}}
A key point in the claim of Theorem~\ref{thm.1} is to determine the hyperparameter $\rho_{a+1}$ that bound the dependency between last layer in subnetwork $f_a:=F^{(a)}$ and first layer in subnetwork $f_b:=F^{(a+1,n)}$ and hyperparameters $\rho_{a+2},\ldots,\rho_n$ that bound dependencies between consecutive layers in $f_b$. Within the experimental results we denote these values as $\rho_{n},\ldots,\rho_{a+1}$, where $\rho_{n}$ corresponds to the last pair of layers in $f_b$. We have devised a novel adversarial training algorithm to determine the $\rho$-values that are sufficient to achieve full-network robustness when utilizing a robust subnetwork $f_a^*$. 

The procedure labeled Algorithm \ref{Algo.1}, assumes that the mutual dependency between the two parts of a network $F^{(n)}$ can be measured by their MI measure. To retrieve baseline results, we first performs non-adversarial training of the whole network with the original dataset, after which the state of the converged network is stored as $(f_a,f_b)$. Adversarial training is then performed to converge on a fully robust network $(f_a^*,f_b^*)$. The adversarial accuracy of $(f_a^*,f_b^*)$ is recorded as $Acc^*$ to serve as a baseline of robustness. 

In the next stage the algorithm runs $T=10$ trials where the non-robust features in subnetwork $f_b$ are restored, to produce the semirobust network $(f_a^*,f_b)$. This semirobust network is then adversarially trained at a learning rate of 0.001 with $f_a^*$ being frozen. The training for a given trial ends when adversarial accuracy $Acc_t^e$ of the resulting model $(f_a^*, \widetilde{f_b})$ is within a small value $k$ of the baseline $Acc^*$. We use this to denote that the network has achieved a full-network robustness equivalent to $(f_a^*,f_b^*)$, with $f_a^*$ having been a robust subnetwork. Following Theorem \ref{thm.1}, this indicates that the dependency values $\rho$ are above the necessary thresholds. Next, the class conditional MI, $I_{j,t} := \sum_y \pi(y)I(f^{(j-1)};f^{(j)}|y)$, between each pair of consecutive layers from $f^{(a)}$ to $f^{(n)}$, is calculated. As the trials progress the largest testing accuracy achieved ($\widetilde{Acc}$) is updated, along with each trials' $I_{j,t}$ values ($\rho_{a+1}$ to $\rho_n$). 
After adversarial training ends, the lowest $\rho$ values are reported among trials which managed to converge with $Acc^*$.

\subsection{Determining Connectivity Thresholds}
In order to investigate how our claims in Theorem~\ref{thm.1} hold in an empirical setting we implement experiments using Algorithm \ref{Algo.1}.
The tests implement the ResNet-18 and WideResNet-34-10 architectures on CIFAR-10 and CIFAR-100, evaluating adversarial accuracy across the FGSM, PGD, and AutoAttack adversarial attacks. As the same fully robust networks are used for $(f_a^*,f_b^*)$, $Acc^*$ is the same for all $f_b$ sizes of a given combination of attack, network, and dataset. Following from Theorem \ref{thm.1} we would expect to see that when $(f_a^*,\widetilde{f}_b)$ achieves robustness (i.e. when $\widetilde{Acc}-Acc^*<k$), then the mutual dependencies between $f_a^*$ and the subsequent layers in the network should be above the threshold defined in (\ref{assumption:A1}). We see in Table~\ref{tab:maintable} that this $f_b$ subnetwork training frequently manages to regain full network robustness within $k$ of $Acc^*$ for a given attack. For this experiment the number of trainable (e.g. convolutional or linear) layers in $f_b$ is tested at both 4 layers and 12 layers to compare the impact of freezing a larger or smaller semirobust $f_a^*$. For this table all robustness was determined by accuracy on test data attacked with AutoAttack. We report the adversarial test accuracies of the fully robust model $(Acc^*)$, the semirobust network $(f_a^*,f_b)$ denoted $Acc_{sr}$, and the network $(f_a^*, \widetilde{f_b})$ denoted $\widetilde{Acc}$. Details of results evaluated on the other included attacks and clean data can be found in the SM.

\begin{algorithm}[H]
    \SetAlgoLined
    Do subsequent regular and adversarial training of $F^{(n)}$ as $(f_a,f_b)$ and $(f_a^*, f_b^*)$ respectively \\
    Store adversarial test accuracy of $(f_a^*, f_b^*)$ as $Acc^*$ \\
    Set k to be as small as possible \\
    Initialize $\rho_{a+1}, \dots, \rho_n = \infty, \dots, \infty$ \\
     \For{$t = $ 1, \dots, T}
     {
     Load $f_b$; freeze $f_a^*$\\
     \For{$e = $ 1, \dots, E}
     {
     Do one epoch of adversarial training of $f_b$ to get $\widetilde{f_b}$ \\
     Store adversarial test accuracy of $(f_a^*, \widetilde{f_b})$ as $Acc^e_t$\\
     \If{$Acc^* - Acc^e_t \leq k$}
     {Break out of epoch loop and store $Acc^e_t$}
     }
     \For{$j = a+1, \dots, n$}
     {
    Compute $I_{j,t} := \sum_y \pi(y)I(f^{(j-1)};f^{(j)}|y)$
    for all consecutive layers in
    $(f_a^*, \widetilde{f_b})$, then store $I_{j,t}$
     }
     }
    $\widetilde{Acc} =$ largest $Acc^e_t$ \\
    $\rho_j =$ smallest $I_{j,t}$ for $j=a+1,\ldots,n$ among trials where $Acc^*-Acc_t^e \leq k$\\
    Report $\rho_{a+1}, \dots, \rho_n$ and $\widetilde{Acc}$\\
    \caption{Learning Hyperparameter $\rho$}
    \label{Algo.1}
\end{algorithm}

\begin{table*}[t]%{\linewidth}
  \centering
    \setlength{\tabcolsep}{5pt} % Default value: 6pt
        \begin{tabular}[t]{@{}llccccccccccc@{}}
        \toprule
        Network & Dataset  & \# $f_b$ layers & ${Acc}^*$ & $Acc_{sr}$ & $\widetilde{Acc}$& Diff. & \# Epochs & $\rho_n$ & $\rho_{n-3}$ & $\rho_{n-7}$ & $\rho_{n-11}$  \\
        \midrule
        \multirow{4}{*}{\shortstack[l]{ResNet-18}} 
         & \multirow{2}{*}{\shortstack[l]{CIFAR-10}} 
             &  4  & 39.45 & 32.12 & 39.57 & 0.12 & 8.2 & 3.31 & 6.16 & - & - \\
             \cline{3-12}
             & & 12  & 39.45 & 16.6 & 40.13 & 0.68 & 8.3 & 3.19 & 5.46 & 7.41 & 7.80 \\
        \cline{2-12}
        &\multirow{2}{*}{\shortstack[l]{CIFAR-100}} 
             & 4 & 17.38 & 14.65 & 18.4 & 1.02 & 1 & 3.51 & 4.66 & - & - \\
             \cline{3-12}
             & & 12 & 17.38 & 1.16 & 17.9 & 0.52 & 4.8 & 3.15 & 4.54 & 4.84 & 5.11 \\
        \midrule
        \multirow{4}{*}{\shortstack[l]{WideResNet-34}} 
         & \multirow{2}{*}{\shortstack[l]{CIFAR-10}} 
             & 4 & 41.66 & 33.85 & 41.76 & 0.10 & 4.7 & 3.70 & 7.43 & - & - \\
             \cline{3-12}
             & & 12 & 41.66 & 28.32 & 43.29 & 1.63 & 1 & 3.48 & 7.65 & 8.36 & 9.05 \\
        \cline{2-12}
        &\multirow{2}{*}{\shortstack[l]{CIFAR-100}} 
            & 4 & 20.32 & 4.68 & 21.34 & 1.02 & 1 & 3.90 & 5.56 & - & - \\
            \cline{3-12}
            & & 12  & 20.32 & 0.07 & 21.5 & 0.88 & 2.3 & 3.73 & 5.35 & 5.54 & 5.63 \\
        \bottomrule
        \end{tabular}    
  \caption{\textbf{Algorithm 1 results evaluated on AutoAttack:} The summarized results of running Algorithm \ref{Algo.1} for ResNet-18 and WideResNet-34-10 for CIFAR-10 and CIFAR-100 are shown. Robust accuracy on test data attacked with AutoAttack was used to determine the values of $Acc^*$, $Acc_{sr}$, and $\widetilde{Acc}$. The difference between $\widetilde{Acc}$ and $Acc^*$, as well as the average number of epochs of finetuning required to reach the baseline robust accuracy are recorded. The lowest $\rho$ values for a subset of the layers in $\widetilde{f}_b$ are reported from amongst the trials which converged with $Acc^*$, where $\rho_n$ corresponds to the last pair of layers in $\widetilde{f}_b$.}\label{tab:maintable}
\end{table*}

\subsection{Effects of Dataset, Network, and Attack on \texorpdfstring{$\rho$}{Rho}:}
\begin{figure*}
\centering
    \includegraphics[width=0.485\columnwidth]{"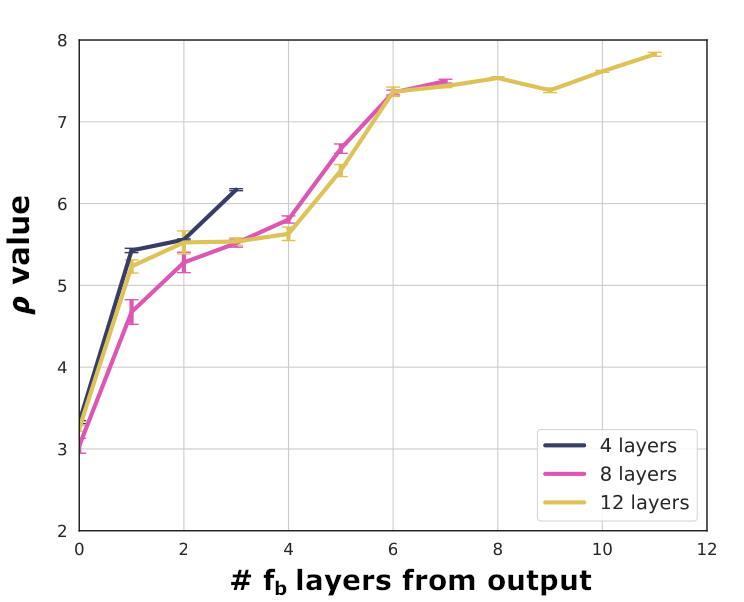"}
    \includegraphics[width=0.485\columnwidth]{"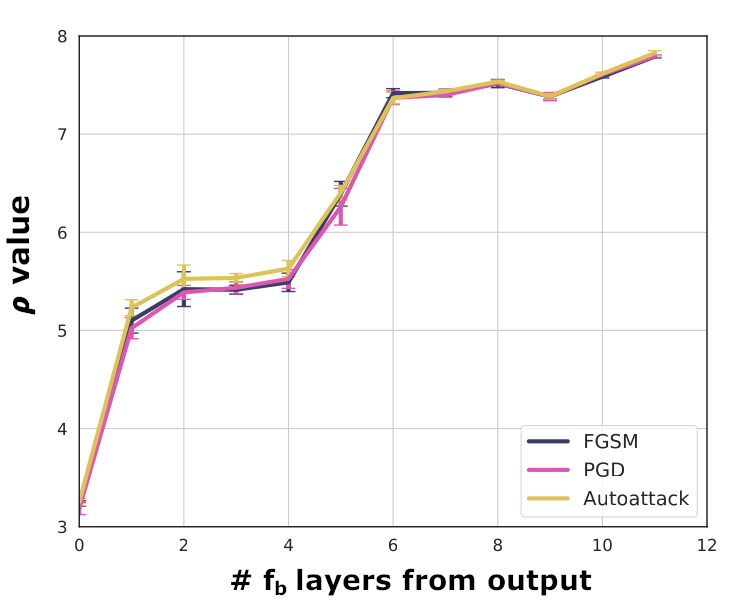"}
\caption{\textbf{$\rho$ value consistency across attacks and subnetwork size:} The observed values of $\rho$ remain largely consistent when varying the size of $f_b$ (left) and when assessing robustness with different adversarial attacks (right). In both cases, these $\rho$ values reflect the layer dependencies in the robust network $(f_a^*,\widetilde{f}_b)$. 
}
\label{fig:consistent}
\end{figure*}

In order to investigate the effects of dataset, network type, and attack type on the observed $\rho$ values, we ran a series of experiments for Algorithm~\ref{Algo.1}, some of which are shown in Figures \ref{fig:consistent} and \ref{fig:variability}, with additional results found in the SM. For all figures, we plot the mean $\rho$ values at each layer among all trials in Algorithm \ref{Algo.1} which reached full network robustness. Figure \ref{fig:consistent} demonstrates the effect of selecting different sizes of $f_b$ during Algorithm \ref{Algo.1}. When $f_b$ comprised of 4 layers slightly higher $\rho$ values were observed in most experiments, while the difference between 8 layers and 12 were minor. For this reason we focus on comparing conditions at 4 and 12 layers for the size of $f_b$. In Figure \ref{fig:consistent} we observe insignificant differences in $\rho$ when evaluating on different adversarial attacks, and as such primarily report our results on AutoAttack, with results on other attacks being provided in the SM. We observe significant differences for $\rho$ between different datasets and networks in Figure \ref{fig:variability}, with CIFAR-100 and ResNet-18 reporting lower $\rho$ values than CIFAR-10 and WideResNet-34-10.  

{\bf Behavior of $\rho$ During Training:}
We run experiments tracking the values of $\rho$ across each layer in $f_b$ during a single trial of Algorithm \ref{Algo.1} in Figure \ref{fig:variability}. As our theory would suggest, the finetuning of $f_b$ with $f_a^*$ remaining frozen progressively increases the connectivity in each layer as the network becomes more robust. However the $\rho$ values are often higher for ($f_a^*,f_b)$ than the subsequent epochs of subnetwork training. Notably the early layers of $\widetilde{f}_b$ have highly consistent $\rho$ values throughout the finetuning.

\subsection{Experimental Analysis}
We observe in our experiments that changes in the dataset and network impact the values of $\rho$. This may reflect the increased number of classes in CIFAR-100 resulting in a more diverse set of learned features, such that fewer features are being strongly relied on for each given class. Similarly, the wider layers in WideResNet may result in redundant features being learned which would potentially increase the observed connectivity values. We show that for deeper layers in the network within $\widetilde{f_b}$, $\rho$ tends to take higher values. This may indicate that deeper layers have a better flow of information which enables $f_{a+1}$ to readily learn to utilize the features in $f_a$. 
We report no clear trends between $\rho$ and any of the attack types used here, which is likely due to using adversarial training with PGD for all experiments, regardless of the attack used for evaluation. 

{\bf Additional Considerations}
Many of the experiments for Algorithm \ref{Algo.1}, including those shown in Table \ref{tab:maintable}, achieve full network robustness with just a few epochs of finetuning. This was seen most consistently in AutoAttack, which suggests that a method utilizing the theory presented here may be able to generalize to different attacks more readily. In this case, we suspect that by reloading the non-robust weights in $f_b$ it helped avoid overfitting to the PGD attack such that $(f_a^*,\widetilde{f}_b)$ was able to easily outperform $(f_a^*,f_b^*)$. 

\begin{figure*}
\centering
    \includegraphics[width=0.485\columnwidth]{"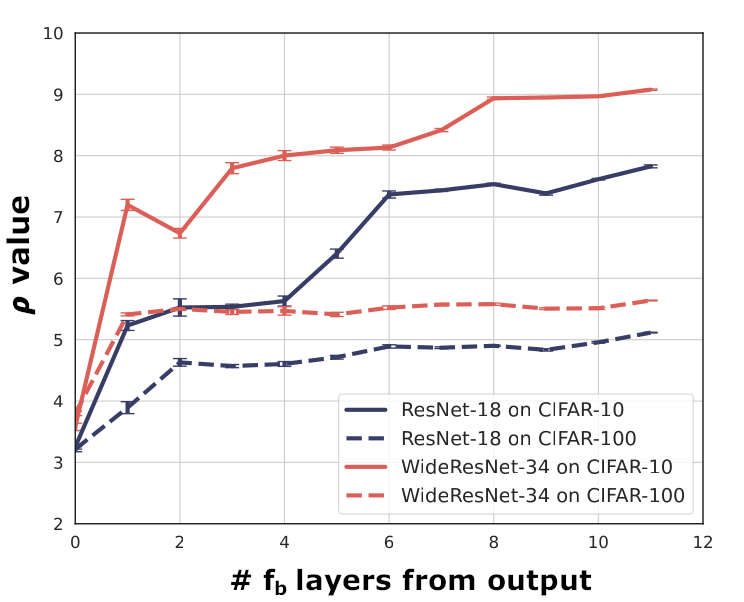"}
    \includegraphics[width=0.485\columnwidth]{"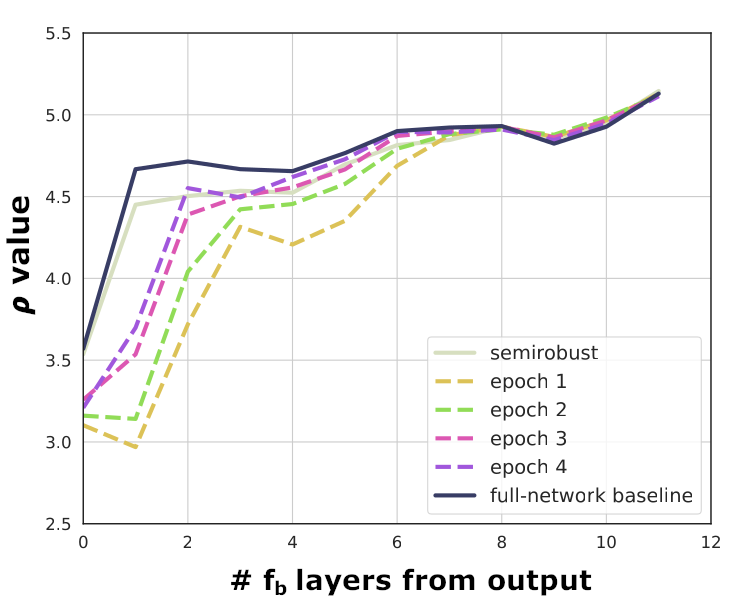"}
\caption{\textbf{Varying $\rho$ values across network changes:} The $\rho$ values varied significantly across differing architectures and datasets (left) and across the epochs of adversarial training (right). Changes to the architecture and data may alter the $\rho$ values by altering the number of filters in a given layer and the number of classes being accounted for, while the variation seen across epochs reflects the increasing accuracy and adversarial robustness of $(f_a^*,\widetilde{f}_b)$.
}
\label{fig:variability}
\end{figure*}

It is also necessary to consider the results from Figure \ref{fig:variability}, which demonstrate that even the normally trained $f_b$ often has comparable connectivity compared to $\widetilde{f}_b$, despite having much lower robustness as measured by adversarial test accuracy. Together this suggests that while connectivity plays a role in adversarial robustness, MI alone may not capture the entire information flow between layers of subnetworks to accurately determine these mutual dependencies.

\section{Related Work}

An important paper that studies adversarial robustness from a theoretical perspective is \cite{ilyas2019adversarial}, which claim that adversarial examples are ``features'' rather than bugs. The authors state that a network's being vulnerable to adversarial attacks ``is a direct result of [its] sensitivity to well-generalizing features in the data''. Specifically, DNNs are learning what they call ``useful, non-robust features'': useful because they help a network improve its accuracy, and non-robust because they are imperceptible to humans and thus not intended to be used for classification. Consequently, a model considers robust features to be about as important as non-robust ones, yet adversarial examples encourage it to rely on only non-robust features. \cite{ilyas2019adversarial} introduces a framework to explain the phenomenon of adversarial vulnerability. Rather than focusing on which features the model is learning, our work's focus is on proving a probabilistic close-form solution to determine the minimal subnetwork which needs to be adversarially trained in order to confer full-network adversarial robustness.

More recently some attention has been given to the adversarially robust subnetworks through methods following the concept from \cite{frankle2018lottery} including \cite{peng2022dynamic} and \cite{fu2021drawing}. Although these works are also interested in robust subnetworks, the focus is often more empirical, or focuses on the robustness of the subnetwork itself as in \cite{guo2022improving}, rather than what we do which is to investigate how other subnetworks can benefit from that semirobustness. Applying the theory outlined here to such methods could provide an interesting avenue for continual learning, where robust subnetworks are sequentially identified and built up over a series of tasks by incorporating the theory behind semirobustness.

\section{Conclusion}
\paragraph{Discussion}
We have introduced here a theoretic framework for the notion of semirobustness, when a part of a network is adversarially robust. The investigation of this characteristic has interesting applications both theoretically and empirically. We prove that if a subnetwork is semirobust and its layers have a sufficient dependency with later layers then the second subnetwork is robust. This has been proven under non-linear dependency (MI)  and linear connectivity between layers in two subnetworks.

As our theory makes no assumptions on how the subnetwork is adversarially trained, it is expected to serve as an orthogonal approach to existing adversarial training methods. We additionally show through our experiments that given a semirobust network where fewer than half of the layers are adversarially robust (as with $(f_a^*,f_b)$ for ResNet-18 when $f_b$ contains the last 12 trainable layers), training the remaining non-robust portion for a small number of epochs can reproduce the robustness of a network which was fully adversarially trained when evaluated on the same attack. In particular the results seen for AutoAttack in Table \ref{tab:maintable} and clean data (shown in the SM) suggest that by only adversarially training a subnetwork under the sufficient mutual dependency constraints there is a potential to limit any trade-off of test accuracy on clean data and to better generalize to different attacks. 

The theory outlined here may help other approaches which are constrained to being applied to the full network, providing a framework to challenge this constraint by finding ways to leverage semirobustness under sufficient conditions for mutual dependency.

\paragraph{Looking ahead} One open question here is how we can determine the complexity of the semirobust subnetwork performance in terms of convergence rate. The answer to this question involves investigating a bound on performance difference as a function of dependency between layers ($\rho$). Another significant question that our experiments pose is how to best measure dependency if MI alone didn't capture all of the necessary information. Lastly, the ability we show in this work to leverage subnetwork robustness of $(f_a^*,f_b)$ to match or outperform the full-network robustness of $(f_a^*,f_b^*)$ could be applied in a future method. The current work requires adversarially training $f_b$ in order to measure $\rho$ thresholds, but applying this concept to a method may require an approach which enforces sufficient dependency between the robust subnetwork and the remainder of the network and provides an interesting future work.

% ---- Bibliography ----

\section{Acknowledgment}
This work has been partially supported by CAREER NSF 5409260 and UMaine AI seed grant; the findings are those of the authors only and do not represent any position of these funding bodies.

\renewcommand{\refname}{References}

\bibliographystyle{IEEEbib}
\bibliography{egbib}

\appendix
\section{Semirobustness Guarantees}

\subsection{Proof of Theorem~1}
First, we show the leftward implication, that if the layers $f^{(j)},f^{(j-1)},\dots,f^{(1)}$ are semirobust, then $F^{(j)}$ is semirobust. This is proved because $F^{(j)}$ is $f^{(j)}(x^{(j-1)})$ where $x^{(j-1)} = f^{(j-1)}\circ\ldots \circ f^{(1)}$. Therefore, if $f^{(j)}$ is semirobust, regardless of whether any of $f^{(j-1)}, \ldots , f^{(1)}$ is semirobust, then $F^{(j)}$ is also semirobust.\\
Next, we show the rightward implication, that if $F^{(j)}$ is semirobust, then $f^{(j)},f^{(j-1)},\dots,f^{(1)}$ are semirobust. If $F^{(j)}$ is semirobust then
\begin{equation}
    \begin{aligned} 
        &\mathbb{E}_{(\mathbf{X},y)\sim D}\left[\inf_{\delta\in S} y \cdot G_j\circ F^{(j)}(\mathbf{X}+\delta)\right]\\ &= \mathbb{E}_{(\mathbf{X},y)\sim D}\left[\inf_{\delta\in S} y \cdot G_j\circ f^{(j)}(\mathbf{X}+\delta)\right]\geq\gamma_j 
    \end{aligned}
\end{equation}
This implies that $f^{(j)}$ is semirobust. Now let $G_{j-1}=G_j\circ f^{(j)}$, then
\begin{equation}
    \begin{aligned}
    &\mathbb{E}_{(\mathbf{X},y)\sim D}\left[\inf_{\delta\in S} y \cdot G_j\circ F^{(j)}(\mathbf{X}+\delta)\right]\\ &=\mathbb{E}_{(\mathbf{X},y)\sim D}\left[\inf_{\delta\in S} y \cdot G_j\circ f^{(j)}\circ F^{(j-1)}(\mathbf{X}+\delta)\right] 
    \\ &=\mathbb{E}_{(\mathbf{X},y)\sim D}\left[\inf_{\delta\in S} y \cdot G_{j-1}\circ F^{(j-1)}(\mathbf{X}+\delta)\right]\\ &=\mathbb{E}_{(\mathbf{X},y)\sim D}\left[\inf_{\delta\in S} y \cdot G_{j-1}\circ f^{(j-1)}(\mathbf{X}+\delta)\right]\geq\gamma_j, 
    \end{aligned}
\end{equation}
This implies that $f^{(j-1)}$ is semirobust. By induction, it's shown that the other layers $f^{(j-2)},\dots,f^{(1)}$ are also semirobust.
%-------------------------------
\subsection{Proof of Lemma~1}
Let $f^{(n-1)}=g\in \mathcal{L}_{n-1}$ and $f^{(n)}=h\in \mathcal{L}_n$. As $f^{(n-1)}$ is semirobust,
\begin{equation}
        \mathbb{E}_{(\mathbf{X},y)\sim D}\left[\inf_{\delta\in S} y\cdot G_{n-1}\circ g(\mathbf{X}+\delta)\right] \geq \gamma_{n-1} 
\end{equation}
and
\begin{equation}
    \label{simplify2}\sum\limits_y \pi(y) I\left(g_\delta;h_\delta|y\right) \geq \rho,    
\end{equation}
and after simplification, we have
\def\bx{\mathbf{x}}
\begin{equation}
    \begin{aligned}\label{eq:2-2}
        \mathbb{E}_{(\mathbf{X},y)\sim D}\left[\inf_{\delta\in S} y\cdot G_{n-1}\circ g(\mathbf{X}+\delta)\right] = \sum_y y \cdot \pi(y) \int \inf_{\delta\in S} D(x|y) \cdot G_{n-1} \circ g(x+\delta) d\mathbf{x} 
    \end{aligned}
\end{equation}

with $\pi(y)$ being the prior of $y$, and $D(x|y)$ being the probability density function of $x$ and $y$.
Let $g_\delta = g(\bx+\delta)\in\mathcal{L}_{n-1}$, with components $g^{(i)}_\delta=g_i(\bx+\delta)$ such that $\bf{x}=(x^{(1)},\ldots,x^{(d)})\in \mathcal{R}^d$, and $g=(g_1,\ldots,g_{m})\in \mathcal{L}_{n-1}$. Note that the multivariate transformation $g_i$ is one to one; hence, the transformation is invertible and can be solved for the equation $x^{(i)}+\delta^{(i)}=g_i^{-1}(g_\delta)$.

Thus, the last line in (\ref{eq:2-2}) equals
\begin{align*}\label{eq:2-3}
 \sum_y y \cdot \pi(y) \int  \inf_{\delta\in S} D(g^{-1}(g_\delta)-\delta|y) \cdot G_{n-1} \circ g_\delta\; |J|\; d g_\delta 
\end{align*}

where $|J|$ denotes the absolute value of the determinant of the Jacobian $J$. In addition, using the probability density for a function of a random variable, we can write:
\begin{equation}
    \begin{aligned}
        \mathbb{E}_{(\mathbf{X},y)\sim D}\left[\inf_{\delta\in S} y\cdot G_{n-1}\circ g(\mathbf{X}+\delta)\right] =  \sum_y y \cdot \pi(y) \int \inf_{\delta\in S} p(g_\delta|y) \cdot G_{n-1} \circ g_\delta\; d g_\delta 
    \end{aligned}
\end{equation}

By simplifying mutual information $I(g_\delta;h_\delta|y)$, using $\log(x)\leq x+1$,
and recalling assumption {\bf B1}:
\begin{equation}\label{SM.1}
    \begin{aligned}
        \sum_y \pi(y) \iint \inf_{\delta\in S} p(g_\delta, h_\delta|y) \left[\frac{p(g_\delta, h_\delta|y)}{p(g_\delta|y)p(h_\delta|y)}\right]\cdot dg_\delta dh_\delta + 1 \geq \rho
    \end{aligned}
\end{equation}

To show that $f^{(n)}$ is $\gamma_n$-semirobust, we prove
\begin{align*}
\gamma_n\leq \mathbb{E}_{(\mathbf{X},y)\sim D}\left[\inf_{\delta\in S} y\cdot h(\mathbf{X}+\delta)\right].
%\leq \inf_{\delta\in S} \sum_y y \cdot \pi(y) \int p(h_\delta|y) \cdot h_\delta dh_\delta
\end{align*}

Set $\gamma_n \leq \gamma_{n-1} + \rho$, hence we need to show that
\begin{align}\label{align-1}
\gamma_{n-1}+\rho \leq
 \sum_y y \cdot \pi(y) \int \inf_{\delta\in S} p(h_\delta|y) \cdot h_\delta dh_\delta
\end{align}

Following (\ref{SM.1}) and semirobustness for $f^{(n-1)}$, the inequality (\ref{align-1}) can be transformed into
\begin{equation} 
\begin{aligned}
\sum\limits_y  \pi(y) \iint \inf_{\delta\in S} p(g_\delta, h_\delta|y)\; \cdot\Big(y \cdot G_{n-1} \circ g_\delta + \frac{p(g_\delta, h_\delta|y)}{p(g_\delta|y)p(h_\delta|y)} - y\cdot h_\delta + 1\Big) \cdot dg_\delta dh_\delta \leq 0
\end{aligned}
\end{equation}

and subsequently into
\begin{equation}
    \begin{aligned}
        \sum\limits_y \pi(y)\mathbb{E}_{p(g_\delta, h_\delta|y)}[\inf_{\delta\in S}y \cdot (G_{n-1} \circ g_\delta - h_\delta)] \leq -(1+U),
    \end{aligned}
\end{equation}

which holds true recalling the assumption {\bf B2}. This concludes the proof of the lemma.
%-------------------------------------------------------

\subsection{Proof of Theorem~2}
To prove Theorem~2, given that
\begin{equation}\label{simplify1}
    \begin{aligned}
    &\mathbb{E}_{(\mathbf{X},y)\sim D}\left[\inf_{\delta\in S} y\cdot G_a\circ h_a(\mathbf{X}+\delta)\right] \geq \gamma_a \;\; \hbox{and}\\ & \sum\limits_y \pi(y) I\left(h_{\delta,a};h_{\delta,a+1}|y\right) \geq \rho_{a+1},
    \end{aligned}
\end{equation}

We need to show from the inequalities above that for $\gamma_{a+1}\leq \gamma_a+\rho_{a+1}$
\begin{equation}\label{simplify3}
\mathbb{E}_{(\mathbf{X},y)\sim D}\left[\inf_{\delta\in S} y\cdot G_{a+1} \circ h_{a+1}(\mathbf{X}+\delta)\right] \geq \gamma_{a+1}\end{equation}

Under assumptions {\bf A1} and {\bf A2} for $j = a + 1$, we can simplify (\ref{simplify1}) and (\ref{simplify3}). We then need to show
\begin{equation}
    \begin{aligned}
        \sum\limits_y \pi(y)\mathbb{E}_{p(g_\delta, h_{\delta,a+1}|y)}[\inf_{\delta\in S}y \; \cdot  (G_a \circ h_{\delta,a} - G_{a+1} \circ h_{\delta,a+1})] \leq -(1+U_{a+1})
    \end{aligned}
\end{equation}

The above holds true recalling the assumption {\bf A2} and $\pi(y)$ being non-negative. Hence, $f^{(a+1)}$ is $\gamma_{a+1}$-semirobust. And because $f_a = F^{(a)}$ is $\gamma_a$-semirobust, then according to Theorem~1, $F^{(a+1)}$ is also $\gamma_{a+1}$-semirobust.

Similarly, since $f_a=F^{(a+1)}$ is $\gamma_{a+1}$-semirobust, and by assumptions {\bf A1} and {\bf A2} for $j=a+2$, it is implied that $f^{(a+2)}$ is $\gamma_{a+2}$-semirobust. Recursively, it can be shown that all layers in $f_b$, i.e. $f^{(a+1)},\ldots,f^{(n)}$, are $\gamma_j$-semirobust for $j=a+1,\ldots,n$ respectively. Then, according to Theorem~1, $f_b$ is $\gamma_b$-semirobust where $\gamma_b\leq \gamma_a+\sum\limits_{j=a+1}^b\rho_j$, proving Theorem~2.

%----------------------
\subsection{Proof of Lemma~2}
\begin{equation}
    \begin{aligned}
        \mathbb{E}_{(\mathbf{X},y)\sim D}\left[\inf_{\delta\in S} y\cdot f^{(n)}(\mathbf{X}+\delta)\right] =
        \mathbb{E}_{(\mathbf{X},y)\sim D}\left[\inf_{\delta\in S} y\cdot \sum_{i=1}^{n-1}\lambda_i^{T}\cdot f^{(i)}(\mathbf{X}+\delta)\right]
    \end{aligned}
\end{equation}

\begin{equation}
    \begin{aligned}
        \sum_{i=1}^{n-1} \mathbb{E}_{(\mathbf{X},y)\sim D}\left[\inf_{\delta\in S} y\cdot \lambda_i^{T}f^{(i)}(\mathbf{X}+\delta)\right] =\sum_{i=1}^{n-1} \mathbb{E}_{(\mathbf{X},y)\sim D}\left[\inf_{\delta\in S} y\cdot \lambda_i^{T} F^{(i)}(\mathbf{X}+\delta)\right] 
    \end{aligned}
\end{equation}
The last equality holds true because the noises are added to the input $\mathbf{X}$ and since in feed-forward network, each layer is a function of the previous layer therefore $f^{(i)}(\mathbf{X}+\delta)=F^{(i)}(\mathbf{X}+\delta)$.\\\\
Next, by letting $G_i=\lambda_i^T$, then we have
\begin{equation}
    \begin{aligned}
        &\mathbb{E}_{(\mathbf{X},y)\sim D}\left[\inf_{\delta\in S} y\cdot f^{(n)}(\mathbf{X}+\delta)\right]\\ &=\sum_{i=1}^{n-1} \mathbb{E}_{(\mathbf{X},y)\sim D}\left[\inf_{\delta\in S} y\cdot G_i \circ F^{(i)}(\mathbf{X}+\delta)\right]\\ &\geq \sum_{i=1}^{n-1} \gamma_i:=\gamma_n.
    \end{aligned}
\end{equation}
The first inequality is true because of Theorem~1. This concludes that $f^{(n)}$ is $\gamma_n$-semirobust.

\subsection{Proof of Theorem~3}
In Theorem~3, note that if $f_b=f^{(n-1)}$ and $f_a=f_{n-1}=F^{(1,n-1)}$, then it turns to Lemma~2. We prove the theorem where $f_b=F^{(n-1,n)}$ and $f_a=F^{(1,n-2)}$, and the general case $f_a$ and $f_b$ can be shown similarly by extension.
Let $G_{b}:\mathcal{L}_b\mapsto \mathcal{Y}$ be a function that maps layer $f_b$ to the output $y$.
Proof of the case where $f_b=F^{(n-1,n)}$ and $f_a=F^{(n-2)}$:
\begin{equation}
    \begin{aligned} 
    &\mathbb{E}_{(\mathbf{X},y)\sim D}\left[\inf_{\delta\in S} y\cdot F^{(n-1,n)}(\mathbf{X}+\delta)\right]\\ & = \mathbb{E}_{(\mathbf{X},y)\sim D}\left[\inf_{\delta\in S} y\cdot f^{(n)}(\mathbf{X}+\delta)\right] 
\end{aligned}
\end{equation}

Given that $f^{(n)}$ is a linear combination of all the other layers, with $\lambda_{in}^T$ mapping $f^{(i)}$ to $y$,
\begin{equation*}
    \begin{aligned}
    =&\mathbb{E}_{(\mathbf{X},y)\sim D}\left[\inf_{\delta\in S} y\cdot \sum_{i=1}^{n-1}\lambda_{in}^{T}\cdot f^{(i)}(\mathbf{X}+\delta)\right]\\ &=\sum_{i=1}^{n-1}\mathbb{E}_{(\mathbf{X},y)\sim D}\left[\inf_{\delta\in S} y\cdot \lambda_{in}^{T}\cdot f^{(i)}(\mathbf{X}+\delta)\right] 
    \end{aligned}
\end{equation*}
\begin{equation}
    \begin{aligned}\label{sum} 
        & =\sum_{i=1}^{n-2}\mathbb{E}_{(\mathbf{X},y)\sim D}\left[\inf_{\delta\in S} y\cdot\lambda_{in}^{T} \cdot f^{(i)}(\mathbf{X}+\delta)\right]\\ & + \mathbb{E}_{(\mathbf{X},y)\sim D}\left[\inf_{\delta\in S} y \cdot \lambda_{n-1(n)}^{T} \cdot f^{(n-1)}(\mathbf{X}+\delta)\right] 
    \end{aligned}    
\end{equation}
Let $G_i = \lambda_{in}^T$, and let $\alpha$ be the second term in (\ref{sum}). Then, using Theorem~1,
\begin{equation}
    \begin{aligned}
         =\sum_{i=1}^{n-2}&\mathbb{E}_{(\mathbf{X},y)\sim D}\left[\inf_{\delta\in S} y\cdot G_i\circ f^{(i)}(\mathbf{X}+\delta)\right] + \alpha \\ &\geq\sum_{i=1}^{n-2} \gamma_i + \alpha  = \gamma_a + \alpha 
    \end{aligned}
\end{equation}
where $\gamma_a=\sum\limits_{i=1}^{n-2} \gamma_i$. Now simplifying $\alpha$, given that $f^{(n-1)}$ is a linear combination of the layers before it, with $\lambda_{i(n-1)}^T$ mapping $f^{(i)}$ to $f^{(n-1)}$:
\begin{equation*}    
        \mathbb{E}_{(\mathbf{X},y)\sim D}[\inf_{\delta\in S} y\cdot \lambda_{n-1(n)}^{T}\cdot f^{(n-1)}(\mathbf{X}+\delta)]\\ 
\end{equation*}
\begin{equation}
\begin{aligned}
 = \mathbb{E}_{(\mathbf{X},y)\sim D}&[\inf_{\delta\in S} y\cdot \lambda_{n-1(n)}^{T}\cdot \sum_{i=1}^{n-2}\lambda_{i(n-1)}^{T}\cdot f^{(i)}(\mathbf{X}+\delta)]
    \end{aligned}
\end{equation}

\begin{equation}
    \begin{aligned} 
        =\sum_{i=1}^{n-2}\mathbb{E}_{(\mathbf{X},y)\sim D}&[\inf_{\delta\in S} y]\cdot \lambda_{n-1(n)}^{T} \cdot \lambda_{i(n-1)}^{T}\cdot f^{(i)}(\mathbf{X}+\delta)]. 
    \end{aligned}
\end{equation}

Let $\tilde{G}_i = \lambda_{n-1(n)}^{T}\cdot \lambda_{i(n-1)}^{T}$. Then, using Theorem~1,

\begin{equation}
\begin{aligned} 
    =\sum_{i=1}^{n-2}\mathbb{E}_{(\mathbf{X},y)\sim D}&\left[\inf_{\delta\in S} y\cdot \tilde{G}_i\circ f^{(i)}(\mathbf{X}+\delta)\right]\\ & \geq\sum_{i=1}^{n-2} \gamma_i = \gamma_a 
\end{aligned}
\end{equation}

With both terms simplified,
$\gamma_a + \alpha \geq \gamma_a + \gamma_a = \gamma_b$.
Therefore, $f_b$ is semirobust. This proof can be extended to any other combination of $f_a$ and $f_b$. Let's show the case where $f_b=F^{(a+1,n)}$ and $f_a=F^{(a)}$:
\begin{equation}
\begin{aligned} 
\mathbb{E}_{(\mathbf{X},y)\sim D}&\left[\inf_{\delta\in S} y\cdot F^{(a+1,n)}(\mathbf{X}+\delta)\right] \\ &= \mathbb{E}_{(\mathbf{X},y)\sim D}\left[\inf_{\delta\in S} y\cdot f^{(n)}(\mathbf{X}+\delta)\right] 
\end{aligned}
\end{equation}
Given that $f^{(n)}$ is a linear combination of all the other layers, with $\lambda_{in}^T$ mapping $f^{(i)}$ to $y$,
\begin{equation*}
\begin{aligned} 
&=\mathbb{E}_{(\mathbf{X},y)\sim D}\left[\inf_{\delta\in S} y\cdot \sum_{i=1}^{n-1}\lambda_{in}^{T}\cdot f^{(i)}(\mathbf{X}+\delta)\right] \\ & =\sum_{i=1}^{n-1}\mathbb{E}_{(\mathbf{X},y)\sim D}\left[\inf_{\delta\in S} y\cdot \lambda_{in}^{T}\cdot f^{(i)}(\mathbf{X}+\delta)\right] 
\end{aligned}
\end{equation*}
\begin{equation}
\begin{aligned}\label{sum2} &=\sum_{i=1}^{a}\mathbb{E}_{(\mathbf{X},y)\sim D}\left[\inf_{\delta\in S} y\cdot \lambda_{in}^{T}\cdot f^{(i)}(\mathbf{X}+\delta)\right]\\ & + \sum_{i=a+1}^{n-1}\mathbb{E}_{(\mathbf{X},y)\sim D}\left[\inf_{\delta\in S} y\cdot \lambda_{in}^{T}\cdot f^{(i)}(\mathbf{X}+\delta)\right] 
\end{aligned}
\end{equation}

Let $G_i = \lambda_{in}^T$, and for $i=a+1,\ldots,n-1$ let
\begin{equation}
 \alpha_i=\mathbb{E}_{(\mathbf{X},y)\sim D}\left[\inf_{\delta\in S} y\cdot \lambda_{in}^{T}\cdot f^{(i)}(\mathbf{X}+\delta)\right].
\end{equation}
Then, by using Theorem~1 again we have,
\begin{equation}
\begin{aligned} &\sum_{i=1}^a\mathbb{E}_{(\mathbf{X},y)\sim D}\left[\inf_{\delta\in S}  y\cdot G_i\circ f^{(i)}(\mathbf{X}+\delta)\right] + \sum_{i=a+1}^{n-1}\alpha_i \\ &\geq\sum_{i=1}^a \gamma_i + \sum_{i=a+1}^{n-1}\alpha_i \\ &= \gamma_a + \sum_{i=a+1}^{n-1}\alpha_i \end{aligned}
\end{equation}
where $\gamma_a=\sum\limits_{i=1}^a \gamma_i$. Next we show that
$
\sum_{i=a+1}^{n-1}\alpha_i\geq \gamma_a\Big((n-1-a)(n-a)\big/2\Big), 
$
and conclude the proof by setting $\gamma_b:=\gamma_a+\gamma_a\Big((n-1-a)(n-a)\big/2\Big).$
Now by the assumption that 
\begin{equation}\label{assumption}
    f^{(i)}=\sum_{\ell=1}^{i-1} \lambda_{\ell i}^T\cdot f^{(\ell)},
\end{equation}
with $\lambda_{\ell i}^T$ mapping $f^{(\ell)}$ to $f^{(i)}$, then simplifying $\alpha_i$ yields
\begin{equation}
\begin{aligned}
 \alpha_i&=\mathbb{E}_{(\mathbf{X},y)\sim D}\left[\inf_{\delta\in S} y\cdot \lambda_{in}^{T}\sum_{\ell=1}^{i-1} \lambda_{\ell i}^T\cdot f^{(\ell)}(\mathbf{X}+\delta)\right]\\ &=\sum_{\ell=1}^{i-1}\mathbb{E}_{(\mathbf{X},y)\sim D}\left[\inf_{\delta\in S} y\cdot \lambda_{in}^{T}\lambda_{\ell i}^T\cdot f^{(\ell)}(\mathbf{X}+\delta)\right]
\end{aligned}
\end{equation}
for $ i=a+1,\ldots,n-1$. Therefore we have
\begin{equation*}
    \begin{aligned}
     &\sum_{i=a+1}^{n-1} \alpha_i=\\ &\sum_{i=a+1}^{n-1}\sum_{\ell=1}^{i-1}\mathbb{E}_{(\mathbf{X},y)\sim D} \left[\inf_{\delta\in S} y\cdot  \lambda_{in}^{T}\cdot  \lambda_{\ell i}^T\cdot f^{(\ell)}(\mathbf{X}+\delta)\right]
    \end{aligned}
\end{equation*}
\begin{equation}
    \begin{aligned}
    = \sum_{i=a+1}^{n-1}\sum_{\ell=1}^{i-1}\mathbb{E}_{(\mathbf{X},y)\sim D}\left[\inf_{\delta\in S} y\cdot G_{\ell n}\circ f^{(\ell)}(\mathbf{X}+\delta)\right]
    \end{aligned}
\end{equation}
   
where $G_{\ell n}:=\lambda_{in}^T\cdot \lambda_{\ell i}^T$. Under the assumption (\ref{assumption}), we know that for $i=a+1,\ldots,n-1$,
\begin{equation}
    \sum_{\ell=1}^a \mathbb{E}_{(\mathbf{X},y)\sim D}\left[\inf_{\delta\in S} y\cdot G_{\ell n}\circ f^{(\ell)}(\mathbf{X}+\delta)\right] \geq \gamma_a,
\end{equation}
Therefore,
\begin{equation} \label{eq:1}
\begin{aligned}
  &\sum_{i=a+1}^{n-1}  \alpha_i\geq (n-1-a)\gamma_a \\ &+\mathbb{E}_{(\mathbf{X},y)\sim D}\left[\inf_{\delta\in S} y\cdot G_{(a+1)n}\circ f^{(a+1)}(\mathbf{X}+\delta)\right] \\ &
   + \sum_{\ell =a+1}^{a+2}\mathbb{E}_{(\mathbf{X},y)\sim D}\left[\inf_{\delta\in S} y\cdot G_{\ell n}\circ f^{(\ell)}(\mathbf{X}+\delta)\right] \\ &+\ldots \\ &+\sum_{\ell =a+1}^{n-2}\mathbb{E}_{(\mathbf{X},y)\sim D}\left[\inf_{\delta\in S} y\cdot G_{\ell n}\circ f^{(\ell)}(\mathbf{X}+\delta)\right]
\end{aligned}
\end{equation}
Without loss of generality, assume that $G_{\ell n}=G_{\ell}$ for all $\ell=a+1,\ldots,n-2$. This simplifies (\ref{eq:1}) as  
\begin{equation}
\begin{aligned}
   &\sum_{i=a+1}^{n-1} \alpha_i\geq  (n-1-a)\gamma_a\\ &+\mathbb{E}_{(\mathbf{X},y)\sim D}\left[\inf_{\delta\in S} y\cdot G_{(a+1)}\circ f^{(a+1)}(\mathbf{X}+\delta)\right] \\ &
   + \sum_{\ell =a+1}^{a+2}\mathbb{E}_{(\mathbf{X},y)\sim D}\left[\inf_{\delta\in S} y\cdot G_{\ell}\circ f^{(\ell)}(\mathbf{X}+\delta)\right]\\ &+\ldots\\ &+\sum_{\ell =a+1}^{n-2}\mathbb{E}_{(\mathbf{X},y)\sim D}\left[\inf_{\delta\in S} y\cdot G_{\ell}\circ f^{(\ell)}(\mathbf{X}+\delta)\right]
\end{aligned}
\end{equation}
Below we show that if $f_a=F^{(1,a)}$ is semi-robust and $f^{(a+1)}$ is a linear combination of layers $f^{(1)},\ldots, f^{(a)}$, with $\lambda_{\ell(a+1)}$ mapping $f^{(\ell)}$ to $f^{(a+1)}$, then $f_{a+1}=F^{(1,a+1)}$ is a semi-robust feature:
\begin{equation*}
    \begin{aligned}
    & \mathbb{E}_{(\mathbf{X},y)\sim D}\left[\inf_{\delta\in S} y\cdot G_{(a+1)}\circ f^{(a+1)}(\mathbf{X}+\delta)\right]\\ &=\mathbb{E}_{(\mathbf{X},y)\sim D}\left[\inf_{\delta\in S} y\cdot G_{(a+1)}\circ \sum_{\ell=1}^a\lambda_{\ell(a+1)}^T f^{(\ell)}(\mathbf{X}+\delta)\right] \\ & =\sum_{\ell=1}^a\mathbb{E}_{(\mathbf{X},y)\sim D}\left[\inf_{\delta\in S} y\cdot G_{(a+1)}\circ \lambda_{\ell(a+1)}^T f^{(\ell)}(\mathbf{X}+\delta)\right]
\end{aligned}
\end{equation*}

\begin{equation}\label{eq:2}
\begin{aligned}=\sum_{\ell=1}^a\mathbb{E}_{(\mathbf{X},y)\sim D}\left[\inf_{\delta\in S} y\cdot G_{\ell}\circ  f^{(\ell)}(\mathbf{X}+\delta)\right]
    \end{aligned}
\end{equation}
where $G_{\ell}=G_{(a+1)}\circ \lambda_{\ell(a+1)}^T$. Since $f_a$ is semi-robust, all layers $f^{(1)},\ldots, f^{(a)}$ are semi-robust. Hence, the right-hand side in (\ref{eq:2}) is greater than or equal to $\sum_{\ell=1}^a \gamma_\ell=\gamma_a$. \\
\\
Consequently, with the same methodology, this can be extended to the following: if $f_a=F^{(1,a)}$ is semi-robust, and $f^{(a+\ell)}$ is a linear combination of layers $f^{(1)},\ldots f^{(a+\ell-1)}$, then $f_{a+\ell}=F^{(1,a+\ell)}$ for $\ell=1,\ldots, n-2-a$ is semi-robust. This implies that the Ineq. (\ref{eq:1}) is lower-bounded by 
\begin{equation}
\begin{aligned}
(n-1-a)\gamma_a+\gamma_a+ \underbrace{\sum_{\ell =a+1}^{a+2} \gamma_a}_{2 \times\gamma_a } +\underbrace{\sum_{\ell =a+1}^{a+3}\gamma_a}_{3\times \gamma_a} +\ldots+\underbrace{\sum_{\ell =a+1}^{n-2}\gamma_a}_{(n-2-a)\times \gamma_a}, 
\end{aligned}
\end{equation}
which is equal to 
\begin{equation}
\begin{aligned}
&(n-1-a)\gamma_a +\gamma_a\sum_{j=1}^{n-2-a} j \\ &= (n-1-a)\gamma_a+\gamma_a\Big((n-2-a)(n-1-a)\Big/2\Big).
\end{aligned}
\end{equation}

This proves that 
$\sum\limits_{i=a+1}^{n-1}\alpha_i\geq \gamma_a\Big((n-1-a)(n-a)\big/2\Big)=\gamma_b $
and completes the proof.

\section{Supplementary Experiments}
\subsection{Linear Implementation}
In order to implement and demonstrate the concept in Theorem 3 regarding a linear setting for semirobustness, we have outlined an algorithm that predicts the labels of each input as a linear combination of the activation values of the layers in the semirobust subnetwork $f_a^*$. Algorithm ~\ref{Algo.2} details how we apply this linearity constraint on the dependency between layers in the subnetworks. For a full network $(f_a^*,f_b^*)$, rather than generating predictions as the outputs of the nonlinear subnetwork $f_b^*$, we instead seek to find a set of parameters $\lambda$ such that we can take a linear combination of the activations within $f_a^*$ to directly predict the labels. By calculating the optimal $\lambda$ values we aim to achieve the baseline robust test accuracy $Acc^*$ on the adversarial data using the frozen $f_a^*$ subnetwork. We outline how we've addressed the problem of finding these optimal $\lambda$ values through a straightforward linear algebra approach to directly solve for them. 

Given a network which has been adversarially trained to be fully robust, denoted as $(f_a^*,f_b^*)$, the approach is to record the outputs from all layers in the frozen $f_a^*$ subnetwork when applied to test data as $\mathbb{F}_a$. We also record the outputs of the final layer in $f_b^*$ which indicate the predicted labels of the fully robust network as $\mathbb{F}^{(n)}$. We then utilize the loss function 
\begin{equation}
    \ell = \sum_{j=a+1}^n ||f^{(j)*} - f^{(j)}||
\end{equation}
to set $\mathbb{F}^{(j)}=\mathbb{F}^{(j)*}$ for $j=n$ and solve for the $\lambda$ values which minimizes the loss in the following steps:
\begin{equation}
\begin{aligned}
    \mathbb{F}_a^* \cdot \mathbb{\lambda}_{j} = \mathbb{F}^{(j)}\\
    \mathbb{F}_a^* \cdot \mathbb{\lambda}_{j} = \mathbb{F}^{(j)*}\\
    {\mathbb{F}_a^*}^{-1} \cdot \mathbb{F}_a^* \cdot \mathbb{\lambda}_{j} = {\mathbb{F}_a^*}^{-1} \cdot \mathbb{F}^{(j)}\\
    \mathbb{I} \cdot \mathbb{\lambda}_{j} = \mathbb{\lambda}_{j} = {\mathbb{F}_a^*}^{-1} \cdot \mathbb{F}^{(j)}    
\end{aligned}
\end{equation}
In this way we calculate the optimal of $\lambda$ through a matrix multiplication between the inverse of the activations in subnetwork $f_a^*$ and the outputs from the fully robust network's final layer. In order to make this approach more tractable we split the problem of solving for $\lambda$ over several batches of data. For each batch, we calculate the activations  ${\mathbb{F}_{a,k}^*}$ and $\mathbb{F}_k^{(j)}$ for the current batch of data before using them to solve for a set of optimal $\lambda_k$ for the given batch. After all batches are finished the average $\lambda$ values are found across all batches to get $\lambda^*$. The baseline test accuracy $Acc^*$ obtained with the nonlinear, fully robust model $(f_a^*,f_b^*)$ is compared against that obtained by applying $f^{(n)}=\sum\limits_{i=1}^{a}{\lambda_{i}^*}^{T}\cdot f^{(i)}$, denoted as $\widetilde{Acc}$. Notably the goal of this algorithm is not to train a linear network capable of generalizing to new data, but rather to demonstrate that for a given semirobust subnetwork $f_a^*$, some set of $\lambda$ can be found which approach or reproduce the same level of accuracy when making predictions on adversarially attacked data, indicating full-network robustness.

%\begin{wrapfigure}{L}{0.5\textwidth}
\begin{algorithm}
Do subsequent regular and adversarial training of $F^{(n)}$ as $(f_a,f_b)$ and $(f_a^*, f_b^*)$ respectively \\
Replace $f_b^*$ with the linear-combination $f^{(n)}=\sum\limits_{i=1}^{a}\lambda_{i}^{T}\cdot f^{(i)}$\\
Initialize list $\lambda$ of batch-wise calculations of $\lambda_k$\\
Initialize list ${\mathbb{F}_{a}^*}$ of batch-wise activation values ${\mathbb{F}_{a,k}^*}$\\
Initialize list $\mathbb{F}^{(j)}$ of batch-wise outputs of $(f_a^*,f_b^*)$\\
\For{each batch = $k$ of testing data}
{
    Get outputs of $(f_a^*,f_b^*)$ as $\mathbb{F}_k^{(j)}$ and append to $\mathbb{F}^{(j)}$\\
    Calculate the $(f_a^*)$ activations ${\mathbb{F}_{a,k}^*}$ and append to ${\mathbb{F}_{a}^*}$\\
    Solve for $\lambda_k={\mathbb{F}_{a,k}^*}^{-1} \cdot \mathbb{F}_k^{(j)}$\\
    Append $\lambda_k$ to list $\lambda$\\
}
Take element-wise average values of $\lambda_k$ across each index of $\lambda$ to get the optimal values $\lambda^*$\\
    % Store test accuracy of $(f_a^*, \widetilde{f_b})$ as $Acc_t^e$
Calculate test accuracy $\widetilde{Acc}$ of outputs $f^{(n)}=\sum\limits_{i=1}^{a}{\lambda_{i}^*}^{T}\cdot f^{(i)}$\\
Report $\widetilde{Acc}$
\caption{Determining Hyperparameter $\lambda$}
\label{Algo.2}
\end{algorithm}

\begin{table}[t]%{\linewidth}
  \centering
    \setlength{\tabcolsep}{3.5pt} % Default value: 6pt
    \begin{tabular}{@{}ccccccccc@{}}
        \toprule
        Attack & \# $f_a^*$ & ${Acc}^*$~($\%$) & $\widetilde{Acc}$~($\%$) & Diff.~($\%$) & $Acc_{rand}$ \\
        \midrule
        \multirow{2}{*}{\shortstack[l]{None}} 
             & 17 & 69.7 & 61.59 & -8.11 & 9.93 \\
             \cline{2-6}
             & 13 & 69.7 & 60.64 & -9.06 & 10.17\\
            \cline{2-6}
             & 9 & 69.7 & 58.47 & -11.23 & 8.58\\
        \midrule
        \multirow{2}{*}{\shortstack[l]{FGSM}} 
             & 17 &  43.08 & 42.91 & -0.17 &  14.67\\
             \cline{2-6}
             & 13 & 43.08 & 42.46 & -0.62 & 12.77\\
            \cline{2-6}
             & 9 & 43.08 & 41.19 & -1.89 & 8.07\\
        \midrule
        \multirow{2}{*}{\shortstack[l]{PGD}} 
             & 17 & 38.63  & 40.41 & 1.78 & 8.91 \\
             \cline{2-6}
             & 13 & 38.63 & 40.13 & 1.5 & 10.1\\
            \cline{2-6}
             & 9 & 38.63 & 39.47 & 0.84 & 6.82\\
        \midrule
        \multirow{2}{*}{\shortstack[l]{AA}} 
             & 17 & 34.11 & 48.5 & 14.39 & 7.12\\
             \cline{2-6}
             & 13 & 34.11 & 48.29 & 14.18 & 9.63\\
            \cline{2-6}
             & 9 & 34.11 & 47.15 & 13.04 & 9.75\\
        \bottomrule
        \end{tabular}    
  \caption{\textbf{Results of Algorithm \ref{Algo.2} on CIFAR-10 with ResNet-18: } This table shows the resulting test accuracies of ResNet-18 on CIFAR-10 under varying attacks. We apply Algorithm ~\ref{Algo.2} across three different sizes of $f_a^*$ for each attack, indicating how many layers are utilized when implementing $\lambda$. The test accuracy of $(f_a^*,f_b^*)$ is reported as $Acc^*$ while the accuracy of $\sum\limits_{i=1}^{a}{\lambda_{i}^*}^{T}\cdot f^{(i)}$ is reported as $\widetilde{Acc}$. The accuracy when using randomly initialized $\lambda$ values is reported for further comparison.} \label{table:lincom-full}
\end{table}

Table~\ref{table:lincom-full} shows the results of applying Algorithm \ref{Algo.2} to ResNet-18 across multiple settings. The reported value of ``\# $f_a$'' indicates the size of $f_a^*$ which is used for making predictions through the linear combination with $\lambda$. For ResNet-18, which has 18 convolutional and linear layers, if the size is 17 then it indicates that only the final linear layer is being replaced. This is the setting which most closely follows Theorem 3, however we also compare the accuracies obtained when using fewer layers in $f_a^*$ to directly predict the label predictions in layer $f^{(n)}$. We report the accuracy $Acc^*$ of the network $(f_a^*,f_b^*)$, which is the regularly calculated test accuracy of the fully robust network. We then report the accuracy when predictions are calculated instead as $\sum\limits_{i=1}^{a}{\lambda_{i}^*}^{T}\cdot f^{(i)}$. We compare the resulting accuracy against the robust baseline and a negative control calculated using random $\lambda$ values.

On weaker attacks, and most notably clean data, predictions made under this linear setting are less accurate than the robust network. This difference decreases when adversarial attacks are applied, and for stronger attacks the predictions made with the linear setting outperform the baseline. For all settings as fewer layers are included in $f_a^*$ the performance of the linear predictions decreases. 

There are intuitive explanations for these observations. In this setting the aim is to demonstrate that the semirobust network $f_a^*$ can be used to gain robustness under a constraint of linearity. We don't make any assumption about the structure of $f_b$, such as only having access to the previous layer, and as such we include all convolutional and linear layers in $f_a^*$ when calculating  $\sum\limits_{i=1}^{a}{\lambda_{i}^*}^{T}\cdot f^{(i)}$. It's expected that replacing the non-linear portion of the network $f_b$ with the linear combination using $\lambda$ would reduce accuracy, which we see for clean data, however the results suggest that for more difficult attacks having access to the outputs of all layers in the subnetwork $f_a^*$ outweighs this detriment. These observations are further supported by the decrease in accuracy as the size of $f_a^*$ decreases, in which case fewer layers' information is available when taking a linear combination resulting in worse performance. 

Ultimately these observations support the concepts in Theorem 3 that the semirobust subnetwork $f_a^*$ can be used in some way under a linear constraint to achieve full-network robustness. This experiment serves first and foremost to demonstrate the underlying theory and has practical limitations. We limit the experiment to ResNet-18 and CIFAR-10 because although directly solving for $\lambda$ in this way highlights that such a $\lambda$ exists, the calculation involved would become intractable on larger datasets where you need to take the inverse of a matrix of all intermediate activations of the network. Additionally, we rely on the outputs of the non-linear layers $f_b^*$ when solving for $\lambda^*$ which requires already having a trained, fully robust network.

\begin{table*}[t]
  \centering
    \setlength{\tabcolsep}{5pt} % Default value: 6pt
        \begin{tabular}[t]{@{}llccccccccccccc@{}}
        \toprule
        Attack & Network & Dataset  & \# $f_b$ & $Acc$ & ${Acc}^*$ & $Acc_{sr}$ & $\widetilde{Acc}$& Diff. & \# Epochs & $\rho_n$ & $\rho_{n-3}$ & $\rho_{n-7}$ & $\rho_{n-11}$  \\
        \midrule
        \multirow{8}{*}{\shortstack[l]{Clean}} 
         & \multirow{4}{*}{\shortstack[l]{RN-18}} 
            & \multirow{2}{*}{\shortstack[l]{CIFAR-10}} 
                 & 4 & 86.4 & 69.7 & 67.7 & \textbf{71.4} & \textbf{1.75} & \textbf{2.0} & 3.42 & 6.27 & - & - \\
                 \cline{4-14}
                 &  & & 12 & 86.4 & 69.7 & 62.5 & 66.3 & -3.38 & - & 2.83 & 5.85 & 6.88 & 7.89 \\
        \cline{3-14}
        &&\multirow{2}{*}{\shortstack[l]{CIFAR-100}} 
                 & 4 & 57.4 & 46.8 & 39.5 & 46.4 & -0.33 & - & 3.61 & 4.61 & - & - \\
             \cline{4-14}
                 &  & & 12 & 57.4 & 46.8 & 31.4 & 42.2 & -4.57 & - & 3.64 & 4.63 & 4.81 & 5.02 \\
         \cline{2-14}
         & \multirow{4}{*}{\shortstack[l]{WRN-34}} 
            & \multirow{2}{*}{\shortstack[l]{CIFAR-10}} 
                 & 4 & 90.9 & 81.3 & 74.2 & \textbf{81.6} & \textbf{0.29} & \textbf{1.0} & 4.20 & 8.76 & - & - \\
                 \cline{4-14}
                 &  & & 12 & 90.9 & 81.3 & 68.9 & 77.7 & -3.62 & - & 3.81 & 8.09 & 8.82 & 8.89 \\
        \cline{3-14}
        &&\multirow{2}{*}{\shortstack[l]{CIFAR-100}} 
              & 4 & 69.0 & 56.2 & 46.4 & 55.9 & -0.35 & - & 3.94 & 5.62 & - & - \\
             \cline{4-14}
              &  & & 12 & 69.0 & 56.2 & 40.0 & 52.0 & -4.19 & - & 3.73 & 5.47 & 5.59 & 5.58 \\
        \midrule
        
        \multirow{8}{*}{\shortstack[l]{FGSM}} 
         &  \multirow{4}{*}{\shortstack[l]{RN-18}} 
             & \multirow{2}{*}{\shortstack[l]{CIFAR-10}} 
                 & 4 & 6.21 & 43.1 & 34.4 & 41.3 & -1.73 & - & 3.28 & 6.29 & - & - \\
                 \cline{4-14}
                 &  & & 12 & 6.21 & 43.1 & 22.5 & 38.1 & -4.96 & - & 2.85 & 5.85 & 6.91 & 7.91 \\
        \cline{3-14}
        &&\multirow{2}{*}{\shortstack[l]{CIFAR-100}} 
                & 4 & 4.37 & 19.8 & 14.9 & 19.2 & -0.55 & - &  3.63 & 4.65 & - & - \\
            \cline{4-14}
                 &  & & 12 & 4.37 & 19.8 & 7.6 & 18.5 & -1.30 & - & 3.62 & 4.65 & 4.83 & 5.03 \\
         \cline{2-14}
         & \multirow{4}{*}{\shortstack[l]{WRN-34}} 
            & \multirow{2}{*}{\shortstack[l]{CIFAR-10}} 
                 & 4 & 4.07 & 50.1 & 39.8 & 49.8 & -0.29 & - & 3.95 & 8.76 & - & - \\
                 \cline{4-14}
                 &  & & 12 & 4.07 & 50.1 & 34.9 & 45.8 & -4.25 & - & 3.78 & 8.04 & 8.84 & 8.90 \\
        \cline{3-14}
        &&\multirow{2}{*}{\shortstack[l]{CIFAR-100}} 
              & 4 & 3.04 & 24.5 & 19.3 & 23.9 & -0.53 & - & 3.95 & 5.63 & - & - \\
             \cline{4-14}
              &  & & 12 & 3.04 & 24.5 & 13.8 & 23.4 & -1.03 & - & 3.72 & 5.46 & 5.60 & 5.58 \\
        \midrule
        
        \multirow{8}{*}{\shortstack[l]{PGD}} 
         &  \multirow{4}{*}{\shortstack[l]{RN-18}} 
             & \multirow{2}{*}{\shortstack[l]{CIFAR-10}} 
                 & 4 & 0 & 38.5 & 29.8 & 36.5 & -1.95 & - & 3.30 & 6.32 & - & - \\
                 \cline{4-14}
                 &  & & 12 & 0 & 38.5 & 16.9 & 33.7 & -4.82 & - & 2.89 & 5.87 & 6.93 & 7.92\\
        \cline{3-14}
       & &\multirow{2}{*}{\shortstack[l]{CIFAR-100}} 
                 & 4 & 0 & 16.0 & 12.3 & 15.9 & -0.08 & - & 3.64 & 4.64 & - & - \\
            \cline{4-14}
                 &  & & 12 & 0 & 16.0 & 5.5 & \textbf{16.3} & \textbf{0.28} & \textbf{9.2} & 3.63 & 4.66 & 4.84 & 5.06 \\
         \cline{2-14}
         & \multirow{4}{*}{\shortstack[l]{WRN-34}} 
            & \multirow{2}{*}{\shortstack[l]{CIFAR-10}} 
                 & 4 & 0 & 43.2 & 34.5 & 42.8 & -0.41 & - & 3.97 & 8.81 & - & - \\
                 \cline{4-14}
                 &  & & 12 & 0 & 43.2 & 30.3 & 41.4 & -1.81 & - & 3.80 & 8.09 & 8.85 & 8.89 \\
        \cline{3-14}
        &&\multirow{2}{*}{\shortstack[l]{CIFAR-100}} 
              & 4 & 0 & 20.4 & 17.0 & 19.6 & -0.82 & - & 3.94 & 5.65 & - & - \\
             \cline{4-14}
              &  & & 12 & 0 & 20.4 & 11.1 & \textbf{20.5} & \textbf{0.14} & \textbf{6.0} & 3.67 & 5.39 & 5.57 & 5.58 \\
        \bottomrule
        \end{tabular}    
  \caption{\textbf{Algorithm 1 results evaluated on varying attacks:} The summarized results of running Algorithm 1 for ResNet-18 and WideResNet-34-10 for CIFAR-10 and CIFAR-100 are shown. Robust accuracy on clean data and data attacked with FGSM or PGD was used to determine the values of $Acc^*$, $Acc_{sr}$, and $\widetilde{Acc}$. Non-robust accuracy $Acc$ is additionally reported. The difference between $\widetilde{Acc}$ and $Acc^*$, as well as the average number of epochs required to reach the baseline robust accuracy are recorded. If no trials converged within 10 epochs, this value is omitted. The lowest $\rho$ values for a subset of the layers in $\widetilde{f}_b$ are reported from amongst the trials which converged with $Acc^*$, where $\rho_n$ corresponds to the last pair of layers in $\widetilde{f}_b$.}\label{tab:alg1sup}
\end{table*}

\subsection{Additional Experiments}
As a supplement to the results from the main paper on AutoAttack, we provide here additional results obtained across clean data, FGSM, and PGD. We report the results of Algorithm 1 of the main paper in Table \ref{tab:alg1sup}. These results span the use of ResNet-18 and WideResNet-34-10 on both CIFAR-10 and CIFAR-100. We ran experiments with either an $f_b$ size of 4 or 12 layers, indicating the number of convolutional or linear layers included in $f_b$. We report the non-robust accuracy $Acc$, the baseline robust accuracy $Acc^*$, the initial semirobust accuracy of $(f_a^*,f_b)$ as $Acc_{sr}$, and the best resulting accuracy of $(f_a^*,\widetilde{f}_b)$ as $\widetilde{Acc}$. We additionally report the difference in accuracy $(\widetilde{Acc}-Acc^*)$, the average number of epochs required to converge, and a subset of the $\rho$ values. If no trials converged for a given setting, the largest $\rho$ values are reported over all trials as an approximation of the thresholds for a fully robust network $(f_a^*,\widetilde{f}_b)$ following the observations in the main paper that adversarial training increases $\rho$ values as accuracy improves.

The results in Table \ref{tab:alg1sup} reflect several of the observations made for AutoAttack in the main paper. Larger semirobust subnetwork $f_a^*$ sizes consistently yielded higher values of $\widetilde{Acc}$, which empirically suggests the intuitive case that when $f_a^*$ constitutes an insufficient portion of the overall network, it may be less capable of conferring robustness to the full network. We do however see a few cases in PGD-attacked data where smaller $f_a^*$ layers resulted in slightly higher values of $\widetilde{Acc}$. Notably, when considering the ability of $(f_a^*,f_b)$ to reach $Acc^*$ through finetuning, compared to the results reported for AutoAttack far fewer settings on FGSM or PGD managed to match or outperform the baseline $Acc^*$. This may be because $(f_a^*,f_b^*)$ was adversarially trained with PGD, leading to a higher level of robustness against it and similar attacks which is harder for the semirobust network $(f_a^*,f_b)$ to reach in the small number of epochs used for each trial. Regarding the reported values of $\rho$, we see the same patterns which have been reported in the main paper, including similar $\rho$ values across different attacks, smaller sizes of $f_b$ yielding higher $\rho$ values for a given layer, and lower values being reported for CIFAR-100 and ResNet-18 relative to CIFAR-10 and WideResNet-34-10. 

\begin{figure*}
\centering
    \includegraphics[width=0.485\columnwidth]{"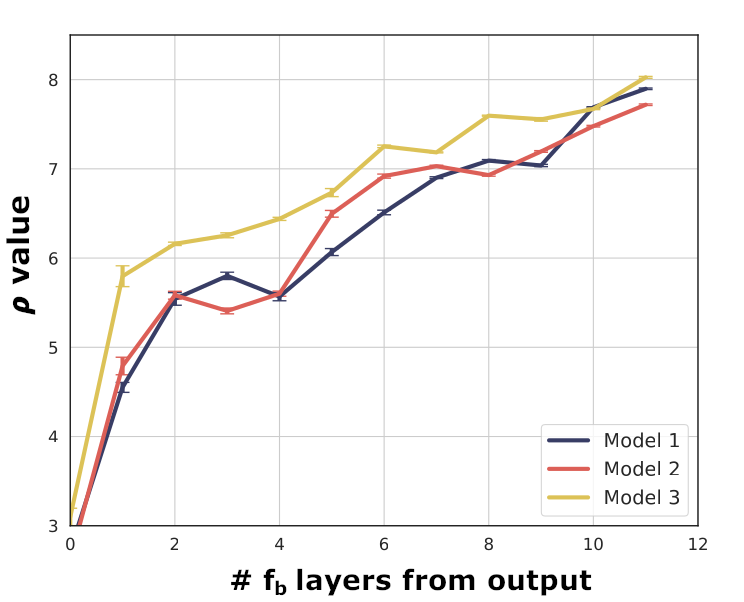"}
    \includegraphics[width=0.485\columnwidth]{"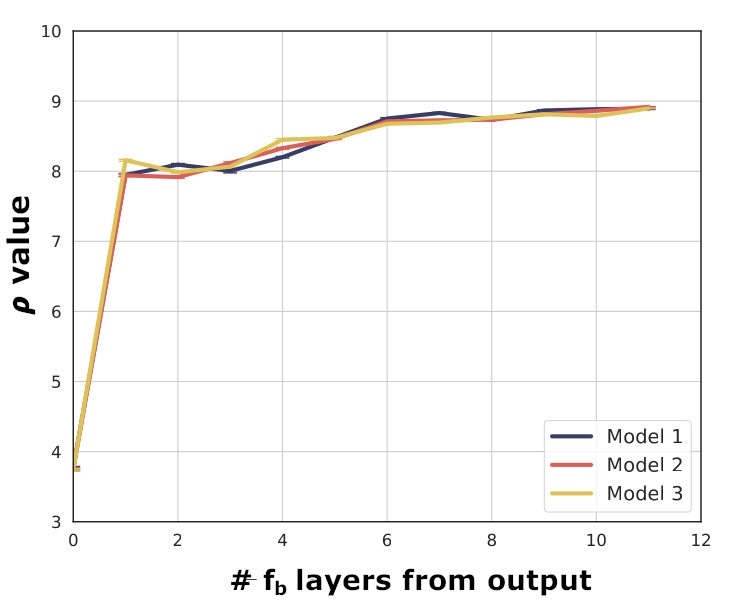"}
\caption{\textbf{Triplicate experiment $\rho$ values:} For ResNet-18 (left) and WideResNet-34-10 (right) on CIFAR-10, the mean $\rho$ values are plotted for the trials of Algorithm 1 across three separately pretrained models.
}
\label{fig:tripl}
\end{figure*}

As Algorithm 1 performs trials over the same initial pretrained model, the results for it are reported among all trials for a given pretrained network. In order to ensure that this didn't lead to erroneous results we ran additional experiments comparing the $\rho$ values obtained for trials run across multiple pretrained models. The results of these experiments for ResNet-18 and WideResNet-34-10 on FGSM-attacked CIFAR-10 are reported in Figure \ref{fig:tripl}. For each setting, triplicate models were initialized and pretrained to yield different $(f_a,f_b)$ and $(f_a^*,f_b^*)$. Trials were then run for each pretrained network, and the average $\rho$ values among trials were plotted for each pretrained network. We observe that when the trials are run on different sets of pretrained weights, resulting $\rho$ values are generally similar. It's necessary to note that significant differences were observed among duplicates of ResNet-18 in Figure \ref{fig:tripl}, however the previously reported trends in $\rho$ values are still seen and the difference between duplicates is much less than that between networks or datasets as demonstrated in Table \ref{tab:alg1sup} and Figure \ref{fig:tripl}.

%\newpage
%\input{Proofs}
\end{document}